\documentclass[journal,11pt,draftcls,onecolumn]{IEEEtran}

\usepackage{hero}
\usepackage{amssymb,amstext,amsmath,amsfonts}
\usepackage{psfrag,epsfig,multimedia}
\usepackage{epstopdf}
\usepackage{graphicx}
\usepackage{turnstile}
\usepackage{subfigure}
\usepackage{enumerate}
\usepackage{tikz}
\tikzstyle{matched}=[circle, draw, fill=black!30,
inner sep=0pt, minimum width=4pt,fill=black!100]

\usepackage{natbib}

\def\ra{\mbox{\rm a}}
\newcommand{\qed}{\mbox{} \hfill $\Box$ }

\def\ra{\mbox{\rm a}}
\def\mN{{\mathbb N}}
\def\mX{{\mathbb X}}
\def\mU{{\mathbb U}}
\def\mY{{\mathbb Y}}

\begin{document}
%


\title{Predictive Correlation Screening: Application to Two-stage Predictor Design in High Dimension}

\author{Hamed Firouzi, \textit{University of Michigan}, Bala Rajaratnam, \textit{Stanford University}, Alfred O. Hero, \textit{University of Michigan}}

\maketitle

\begin{abstract}

We introduce a new approach to variable selection, called Predictive
Correlation Screening, for predictor design.
Predictive Correlation Screening (PCS) implements false positive control
on the selected variables, is well suited to small sample sizes, and
is scalable to high dimensions. We establish asymptotic bounds
for Familywise Error Rate (FWER), and resultant mean square
error of a linear predictor on the selected variables. We apply Predictive
Correlation Screening to the following two-stage predictor design problem.
An experimenter wants to learn a multivariate predictor of gene expressions
based on successive biological samples assayed on mRNA arrays. She assays
the whole genome on a few samples and from these assays she selects a small number of
variables using Predictive Correlation Screening.
To reduce assay cost, she subsequently assays only the selected variables on the remaining samples, to learn the predictor coefficients.
We show superiority of Predictive Correlation Screening relative to LASSO and correlation learning (sometimes popularly referred to in the literature as marginal regression or simple thresholding) in terms of performance and computational complexity.

\end{abstract}



%

\section{Introduction}
\let\thefootnote\relax\footnote{This research was supported in part by AFOSR grant FA9550-13-1-0043.}
Consider the problem of under-determined multivariate linear regression in which training data $\{ \bY_i, X_{i1},...,X_{ip} \}_{i=1}^n$ is given and a linear estimate of
the $q$-dimensional response vector $\bY_i$, $1
\leq i \leq n < p$, is desired:
\begin{equation}
\bY_i= \ba_1X_{i1}+\ldots+\ba_pX_{ip}+ \epsilon_i,~1 \leq i \leq n,
\end{equation}
where $X_{ij}$ is the $i$th sample of regressor varialbe (covariate) $X_j$, $\bY_i$ is a vector of response variables,
and $\ba_j$ is the $q$-dimensional vector of regression coefficients
corresponding to $X_j$, $1 \leq i \leq n, 1 \leq j \leq p$. There
are many applications in which the number $p$ of regressor variables is larger than the number $n$ of samples. Such applications arise in text processing of internet documents, gene expression array analysis, combinatorial
chemistry, and others~\citep{guyon2003introduction}.
In this $p\gg n$ situation training a linear predictor becomes difficult due to rank deficient normal equations, overfitting errors, and high computation complexity. Many penalized regression methods have been proposed to deal with this situation, including: LASSO; elastic net; and group LASSO \citep{guyon2003introduction,tibshirani1996regression,efron2004least,buehlmann2006boosting,
yuan2005model,friedman2001elements, buhlmann2011statistics}.  These methods  perform variable selection by minimizing a penalized mean squared error prediction criterion over all the training data. The main drawback of these methods is their high computation requirements for large $p$. In this paper we propose a highly scalable approach to under-determined multivariate regression called Predictive Correlation Screening (PCS).

Like recently introduced correlation screening methods \citep{hero2011large,hero2012hub} PCS screens for connected variables in a correlation graph. However,  unlike these correlation screening methods, PCS screens for connectivity in a bipartite graph between the regressor  variables $\{X_{1}, \ldots, X_{p}\}$ and the response variables $\{Y_{1}, \ldots, Y_{q}\}$. An edge exists in the bipartite graph between regressor variable $j$ and response variable $k$ if the thresholded min-norm regression coefficient matrix $\bA = [ \ba_1, \ldots, \ba_p ]$ has a non-zero $kj$ entry.  When the $j$-th column of this thresholded matrix is identically zero the $j$-th regressor variable is thrown out.

PCS differs from correlation learning, also called marginal regression, simple thresholding, and sure independence screening  \citep{genovese2012comparison,fan2008sure}, wherein the simple sample cross-correlation matrix between the response variables and the regressor variables is thresholded. Correlation learning does not account for the correlation between regressor variables, which enters into PCS through the pseudo-inverse correlation matrix - a quantity that introduces little additional computational complexity for small $n$.

\begin{figure} [ht]
\centering
\includegraphics[width=3.0in]{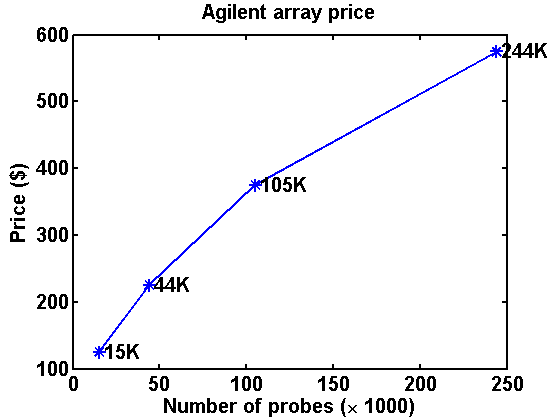}
\caption{Pricing per slide for Agilent Custom Micorarrays G2309F, G2513F, G4503A, G4502A (Feb 2013). The cost increases as a function of probeset size. Source: BMC Genomics and RNA Profiling Core.} \label{fig:Agilent}
\end{figure}

To illustrate our method of PCS we apply it to a two-stage sequential design problem that is relevant to applications where the cost of samples increases with $p$. This is true, for example, with gene microarray experiments: a high throughput ``full genome" gene chip with $p=40,000$ gene probes can be significantly more costly than a smaller assay that tests fewer than $p=15,000$ gene probes (see Fig. \ref{fig:Agilent}).  In this situation a sensible cost-effective approach would be to use a two-stage procedure: first select a smaller number of variables on a few expensive high throughput  samples and then construct the predictor on additional cheaper low throughput samples. The cheaper samples assay only those variables selected in the first stage.

Specifically, we apply PCS to select variables in the first stage of the two-stage procedure. While bearing some similarities, our two-stage PCS approach differs from the many multi-stage adaptive support recovery methods that have been collectively called distilled sensing \citep{haupt2011distilled} in the compressive sensing literature. Like two-stage PCS, distilled sensing (DS) performs initial stage thresholding in order to reduce the number of measured variables in the second stage. However, in distilled sensing   the objective is to recover a few variables with high mean amplitudes from a larger set of initially  measured regressor variables. In contrast, two-stage PCS  seeks to recover a few variables that are strongly  predictive of a response variable from a large number of initially  measured regressor variables and response variables. Furthermore, unlike in DS, in two-stage PCS the final predictor uses all the information on selected variables collected during both stages.

We establish the following theoretical results on PCS and on the two-stage application of PCS. First, we establish Poisson-like limit theorem for the number of variables that pass the PCS screen. This gives a Poisson approximation to the probability of false discoveries that is accurate for small $n$ and large $p$. The Poisson-like limit theorem also specifies a phase transition threshold for the false discovery probability.
Second, with $n$, the number of samples in the first stage, and $t$, the total number of samples, we establish that $n$ needs only be of order $\log(p)$ for two-stage PCS to succeed  with high probability in recovering the support set of the optimal OLS predictor. Third, given a cost-per-sample that is linear in the number of assayed variables, we show that the optimal value of  $n$ is on the order of $\log(t)$. These three results are analogous to theory for correlation screening \citep{hero2011large,hero2012hub}, support recovery for multivariate lasso \citep{obozinski2008high}, and optimal exploration vs exploitation allocation in multi-armed bandits \citep{audibert2007tuning}.

The paper is organized as follows. Section \ref{sec:Prelim} defines the under-determined multivariate regression problem. Section \ref{sec:Theoretical} gives the Poisson-like asymptotic theorem for the thresholded regression coefficient matrix.
Section \ref{sec:Predictive} defines the PCS procedure and associated p-values. Section \ref{sec:two-stage} defines the two-stage PCS and prediction algorithm. Section \ref{sec:Asymptotic} gives theorems on support recovery and optimal sample allocation to the first stage of the two-stage algorithm. Section \ref{sec:Simulations} presents simulation results and an application to symptom prediction from gene expression data.

\section{Under-determined multivariate regression problem}
\label{sec:Prelim}

Assume $\bX=[X_{1},\ldots, X_{p}]$ and
$\bY=[Y_{1},\ldots, Y_{q}]$ are random vectors of regressor and response variables, from which $n$ observations are available. We
represent the $n \times p$ and $n \times q$ data matrices as
$\mathbb X$ and $\mathbb Y$, respectively. We assume
that the vector $\bX$ has an elliptically contoured density
with mean $\mathbf \mu_x$ and non-singular $p\times p$ covariance
matrix $\mathbf \Sigma_x$, i.e. the probability density function is of the form $f_{\bX}(\bx)=
g\left((\bx-\mathbf \mu_x)^T {\mathbf \Sigma_x}^{-1} (\bx-\mathbf
\mu_x)\right)$, in which $g$ is a non-negative integrable
function. Similarly, the vector $\bY$, is assumed to follow an
elliptically contoured density with mean $\mathbf \mu_y$ and
non-singular $q\times q$ covariance matrix $\mathbf \Sigma_y$.
We assume that the joint density function of $\bX$ and $\bY$ is bounded and differentiable. Denote the $p \times q$ population cross covariance matrix
between $\bX$ and $\bY$ by $\mathbf \Sigma_{xy}$.

The $p \times p$ sample covariance matrix $\bS$ for data $\mathbb
X$ is defined as: \be \bS=\frac{1}{n-1} \sum_{i=1}^n
(\bX_{(i)}-\ol{\bX})^T(\bX_{(i)}-\ol{\bX}), \label{sampcov} \ee
where $\bX_{(i)}$ is the $i$th row of data matrix $\mathbb X$, and
$\ol{\bX}$ is the vector average of all $n$ rows of $\mathbb X$.

Consider the $n \times (p+q)$ concatenated matrix $\mathbb
Z=[\mathbb X, \mathbb Y]$. The sample cross covariance matrix
$\bS^{yx}$ is defined as the lower left $q \times p$ block of the
$(p+q) \times (p+q)$ sample covariance matrix obtained by
(\ref{sampcov}) using $\mathbb Z$ as the data matrix instead of
$\mathbb X$.

Assume that $p \gg n$. We define the ordinary least squares (OLS) estimator of $\bY$ given $\bX$ as the min-norm solution of the underdetermined least squares regression problem \be \min_{\bB} \|\mathbb Y^T-\bB \mathbb X^T\|_F^2, \label{eq:OLS1}\ee
where $\| \bA \|_F$ represents the Frobenius norm of matrix $\bA$. The min-norm solution to \eqref{eq:OLS1} is the $q\times p$ matrix of regression coefficients
\be \bB=\bS^{yx}(\bS^x)^\dagger, \ee
where $\bA^\dagger$ denotes the Moore-Penrose pseudo-inverse of matrix $\bA$. If the $i$th  column of $\bB$ is zero then the $i$th variable is not included in the OLS estimator. This is the main motivation for the proposed partial correlation screening procedure.

The PCS procedure for variable selection is based on the U-score
representation of the correlation matrices. It is easily shown that
there exist matrices $\mU^x$ and $\mU^y$
of dimensions $(n-1) \times p$ and $(n-1) \times q$ respectively, such that the columns of $\mU^x$ and $\mU^y$ lie on the $(n-2)$-dimensional unit sphere $S_{n-2}$ in $\mathbb R^{n-1}$ and the following representations hold~\citep{hero2012hub}:
\begin{equation}
\bS^{yx}=\bD^{\frac{1}{2}}_{\bS^y}((\mU^y)^T\mU^x)\bD_{\bS^x}^{\frac{1}{2}},
\label{eq:Syx}
\end{equation}
and:
\begin{equation}
(\bS^x)^{\dagger}=\bD_{\bS^x}^{-\frac{1}{2}}((\mU^x)^T(\mU^x(\mU^x)^T)^{-2}\mU^x)\bD_{\bS^x}^{-\frac{1}{2}},
\label{eq:Sx}
\end{equation}
where $\bD_{\bM}$ denotes the diagonal matrix obtained by zeroing out the off-diagonals of matrix $\bM$. Note that $\mU^x$ and $\mU^y$ are constructed from data
matrices $\mX$ and $\mY$, respectively.

Throughout this paper, we assume the data matrices $\mX$ and $\mY$
have been normalized in such a way that the sample variance of each
variable $X_i$ and $Y_j$ is equal to 1 for  $1 \leq i \leq p$ and
$1 \leq j \leq q$. This simplifies the representations \eqref{eq:Syx} and \eqref{eq:Sx} to $\bS^{yx}=(\mU^y)^T\mU^x$ and $(\bS^x)^{\dagger}=(\mU^x)^T(\mU^x(\mU^x)^T)^{-2}\mU^x$. Using these representations, one can write: \be
\hat{\bY}=\bS^{yx}(\bS^x)^{\dagger} \bX =
(\mU^y)^T(\mU^x(\mU^x)^T)^{-1}\mU^x \bX. \ee Defining
$\tilde{\mU}^x=(\mU^x(\mU^x)^T)^{-1}\mU^x
\bD_{(\mU^x)^T(\mU^x(\mU^x)^T)^{-2}\mU^x}^{-\frac{1}{2}}$, we
have: \be \hat{\bY}=(\mU^y)^T \tilde{\mU}^x
\bD_{(\mU^x)^T(\mU^x(\mU^x)^T)^{-2}\mU^x}^{\frac{1}{2}} \bX \\=
(\bH^{xy})^T
\bD_{(\mU^x)^T(\mU^x(\mU^x)^T)^{-2}\mU^x}^{\frac{1}{2}} \bX, \ee
where \be \bH^{xy} =  (\tilde{\mU}^x)^T \mU^y. \label{Hxy} \ee Note
that the columns of matrix $\tilde{\mU}^x$ lie on $S_{n-2}$.
This can simply be verified by the fact that diagonal entries of
the $p \times p$ matrix $(\tilde{\mU}^x)^T \tilde{\mU}^x$ are
equal to one.

The U-score representations of
covariance matrices completely specify the regression coefficient matrix $\bS^{yx}(\bS^x)^{\dagger}$.

We define variable selection by discovering columns of
the matrix \eqref{eq:CoefMat} that are not close to zero. The expected number of discoveries will play an important role in the theory of false discoveries, discussed below.

From Sec. \ref{sec:Prelim} we obtain a U-score representation of the regression coefficient matrix: \be
\bS^{yx}(\bS^x)^{\dagger} = (\bH^{xy})^T
\bD_{(\mU^x)^T(\mU^x(\mU^x)^T)^{-2}\mU^x}^{\frac{1}{2}}. \label{eq:CoefMat}\ee Under the condition that $\bD_{(\mU^x)^T(\mU^x(\mU^x)^T)^{-2}\mU^x}$ has non-zero diagonal
entries, the $i$th column of $\bS^{yx}(\bS^x)^{\dagger}$ is a
zero vector if and only if the $i$th row of $\bH^{xy}$ is a zero
vector, for $1 \leq i \leq p$. This motivates screening for
non-zero rows of the matrix $\bH^{xy}$ instead of columns of
$\bS^{yx}(\bS^x)^{\dagger}$.

Fix an integer $\delta \in \{1,2,\cdots,p \}$ and a real number
$\rho \in [0,1]$. For each $1 \leq i \leq p$, we call $i$ a
discovery at degree threshold $\delta$ and correlation threshold
$\rho$ if there are at least $\delta$ entries in $i$th row of
$\bH^{xy}$ of magnitude at least $\rho$. Note that this
definition can be generalized to an arbitrary matrix of the form
$(\mathbb{U}^x)^T \mathbb{U}^y$ where $\mathbb{U}^x$ and
$\mathbb{U}^y$ are matrices whose columns lie on $S_{n-2}$.
For a general matrix of the form $(\mathbb{U}^x)^T \mathbb{U}^y$ we
represent the number of discoveries at degree level $\delta$
and threshold level $\rho$ as $N_{\delta, \rho}^{xy}$.

\section{Asymptotic theory}
\label{sec:Theoretical}

The following notations are necessary for the 
propositions in this section. We denote the surface area of the $(n-2)$-dimensional unit sphere $S_{n-2}$ in $\mathbb R^{n-1}$ by $\ra_n$. Assume that $\bU, \bV$ are two independent and
uniformly distributed random vectors on $S_{n-2}$. For a threshold
$\rho \in [0,1]$, let $r=\sqrt{2(1-\rho)}$. $P_0$ is then defined as the
probability that either $\|\bU-\bV\|_2 \leq r$ or $\|\bU+\bV\|_2 \leq r$. $P_0$ can be computed using the formula for the area of spherical caps on $S_{n-2}$ \citep{hero2012hub}.

Define the index set ${\mathcal C}$ as: \be {\mathcal C}=\{(i_0,i_1,
\ldots, i_\delta): \nonumber \\ 1 \le i_0 \le p, 1 \le i_1< \ldots < i_\delta
\le q \}. \label{indexset}\ee
For arbitrary joint density
$f_{\bU_0,\ldots,\bU_\delta}(\bu_0,\ldots,\bu_\delta)$ defined on the Cartesian product
$S_{n-2}^{\delta+1}= S_{n-2} \times \cdots \times S_{n-2}$,  define
$\ol{f_{\bU_{\bullet}^x,\bU_{\ast_1}^y, \ldots,
\bU_{\ast_{\delta}}^y}} (\bfu_0,\bfu_1, \ldots, \bfu_{\delta})$ as
the average of \be f_{\bU_{\vec{i}}}(s_0 \bfu_0,s_1 \bfu_1, \ldots,
s_{\delta} \bfu_{\delta})= \nonumber \\ f_{\bU_{i_0}^x,\bU_{i_1}^y,\ldots,
\bU_{i_\delta}^y}(s_0 \bfu_0,s_1 \bfu_1, \ldots, s_{\delta}
\bfu_{\delta}), \ee for all $\vec{i} = (i_0,i_1, \ldots, i_\delta) \in
{\mathcal C}$ and $s_j \in \{-1,1\}, 0 \le j \le \delta$.

In the following propositions, $k$ represents an upper bound on the number of non-zero entries
in any row or column of covariance matrix $\mathbf \Sigma_x$ or cross covariance
matrix $\mathbf \Sigma_{xy}$. We define $\| \Delta^{xy}_{p,q,n,k,\delta}\|_1 =|\mathcal C|^{-1} \sum_{\vec{i} \in \mathcal C} \Delta^{xy}_{p,q,n,k, \delta}(\vec{i})$, the
average dependency coefficient, as the average of
\be
\Delta_{p,q,n,k,\delta}^{xy}(\vec{i})= \left
\|(f_{\bU_{\vec{i}}|\bU_{A_k(i_{0})}}-f_{\bU_{\vec{i}}})/f_{\bU_{\vec{i}}}
\right\|_{\infty}, \ee in which $A_k(i_{0})$ is defined as the
set complement of the union of indices of non-zero elements of the $i_0$-th column of $\mathbf \Sigma_{yx} \mathbf \Sigma_x^{-1}$. Finally, the function $J$ of the joint
density $f_{\bU_0,\ldots,\bU_\delta}(\bu_0,\ldots,\bu_\delta)$ is
defined as: \be J(f_{\bU_0,\ldots,\bU_\delta}) =
|S_{n-2}|^{\delta} \int_{S_{n-2}}
f_{\bU_0,\ldots,\bU_\delta}(\bu, \ldots,\bu) d\bu. \label{Jdef} \ee

The following proposition gives an asymptotic expression for the
number of discoveries in a matrix of the form $(\mathbb{U}^x)^T
\mathbb{U}^y$, as $p \rightarrow \infty$, for fixed $n$. Also it
states that, under certain assumptions, the probability of having at
least one discovery converges to a given limit. This limit is
equal to the probability that a certain Poisson random variable $N^*_{\delta, \rho_p}$ with rate equal to $\lim_{p\rightarrow \infty} E[N^{xy}_{\delta, \rho_p}]$ takes a non-zero value, i.e. it satisfies: $N^*_{\delta, \rho_p}>0$.

\begin{propositions}
Let $\mathbb{U}^x=[\bU_1^x, \bU_2^x,...,\bU_p^x]$ and $\mathbb{U}^y=
[\bU_1^y, \bU_2^y,...,\bU_q^y]$ be $(n-1)\times p$ and $(n-1) \times q$
random matrices respectively, with $\bU_i^x, \bU_j^y \in S_{n-2}$ for
$1 \leq i \leq p, 1 \leq j \leq q$. Fix integers $\delta \ge 1$
and $n>2$. Assume that the joint density of any subset of
$\{\bU_1^x,...\bU_p^x, \bU_1^y,...,\bU_q^y\}$ is bounded and
differentiable. Let $\{\rho_p\}_p$ be a sequence in $[0,1]$ such
that $\rho_p \rightarrow 1$ as $p\rightarrow \infty$ and
$p^{\frac{1}{\delta}}q(1-\rho_p^2)^{\frac{(n-2)}{2}} \rightarrow
e_{n,\delta}$. Then, \be \nonumber \lim_{p\rightarrow \infty}
E[N^{xy}_{\delta, \rho_p}]= \lim_{p\rightarrow \infty}
\xi_{p,q,n,\delta,\rho_p} J(\overline{f_{\bU_{*}^x,\bU_{\bullet
1}^y,...,\bU_{\bullet \delta}^y}}) \\= \kappa_{n,\delta}
\lim_{p\rightarrow \infty} J(\overline{f_{\bU_{*}^x,\bU_{\bullet
1}^y,...,\bU_{\bullet \delta}^y}}), \label{prop1first} \ee where $
\xi_{p,q,n,\delta,\rho_p}=p{q \choose \delta} P_0^{\delta}
\label{xidef}$ and $ \kappa_{n,\delta}=\left(e_{n,\delta}
\ra_n/(n-2)\right)^\delta/\delta! $.\\
Assume also that $k=o((p^{\frac{1}{\delta}}q)^{1/(\delta+1)})$ and
that the average dependency coefficient satisfies \\ $\lim_{p\rightarrow \infty}
\| \Delta^{xy}_{p,q,n,k,\delta}\|_1=0$. Then:
\be p(N_{\delta,
\rho_p}^{xy}>0) \rightarrow 1-\exp(-\Lambda^{xy}_{\delta}),
\label{eq:Poissonconvcross}\ee
with \be
\Lambda_{\delta}^{xy}= \lim_{p\rightarrow \infty}E[N_{\delta, \rho_p}^{xy}].
\label{eq:Poissonconvcross2}
\ee
\label{Prop1}
\end{propositions}
\noindent{\it Proof of Proposition \ref{Prop1}:} See appendix.

The following proposition states that when the rows of data
matrices $\mX$ and $\mY$ are i.i.d. elliptically distributed with
block sparse covariance matrices, the rate \eqref{prop1first} in Proposition 1 becomes independent of $\Sigma_x$ and $\Sigma_{xy}$. Specifically, the $(\delta+1)$-fold
average $J(\ol{f_{\bU_{\bullet}^x,\bU_{\ast_1}^y, \ldots,
\bU_{\ast_{\delta}}^y}})$ converges to $1$ while the average
dependency coefficient $\| \Delta^{xy}_{p,q,n,k,\delta}\|_1$ goes to $0$, as $p \rightarrow \infty$. 
This proposition will play an an important role
in identifying phase transitions and in approximating $p$-values. 
\begin{propositions}
Assume the hypotheses of Prop. 1 are satisfied. In addition assume that the rows of data
matrices $\mX$ and $\mY$ are i.i.d. elliptically distributed with
block sparse covariance and cross covariance matrices $\mathbf \Sigma_{x}$ and $\mathbf \Sigma_{xy}$. Then $\Lambda_{\delta}^{xy}$ in the limit \eqref{eq:Poissonconvcross2} in Prop. \ref{Prop1} is equal to the constant $\kappa_{n,\delta}$ given in \eqref{prop1first}. Moreover, $\tilde{\mathbb U}_x \approx \mathbb U_x$.
\label{Prop2}
\end{propositions}
\noindent{\it Proof of Proposition \ref{Prop2}:} See appendix.
\section{Predictive Correlation Screening}
\label{sec:Predictive}
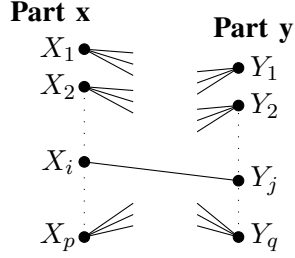
\begin{figure}
\begin{center}
\begin{tikzpicture}[thin]

\node at (-0.1,-0.5) {$\textbf{Part x}$};
\node at (0,-1) {$X_1$};
\node at (0,-1.5) {$X_2$};
\node at (0,-2.5) {$X_i$};
\node at (0,-3.5) {$X_p$};
\node at (2.6,-0.75) {$\textbf{Part y}$};
\node at (2.75,-1.25) {$Y_1$};
\node at (2.75,-1.75) {$Y_2$};
\node at (2.75,-2.75) {$Y_j$};
\node at (2.75,-3.5) {$Y_q$};

\node at (0.35,-1) [matched]{};
\node at (0.35,-1.5) [matched]{};
\node at (0.35,-2.5) [matched]{};
\node at (0.35,-3.5) [matched]{};
\node at (2.4,-1.25) [matched]{};
\node at (2.4,-1.75) [matched]{};
\node at (2.4,-2.75) [matched]{};
\node at (2.4,-3.5) [matched]{};

\draw (0.35,-1) -- (1,-1.05);
\draw (0.35,-1) -- (1,-1.2);
\draw (0.35,-1) -- (1,-1.35);
\draw (0.35,-1.5) -- (1,-1.55);
\draw (0.35,-1.5) -- (1,-1.7);
\draw (0.35,-1.5) -- (1,-1.85);
\draw (0.35,-3.5) -- (1,-3.05);
\draw (0.35,-3.5) -- (1,-3.2);
\draw (0.35,-3.5) -- (1,-3.35);

\draw (2.4,-1.25) -- (1.85,-1.30);
\draw (2.4,-1.25) -- (1.85,-1.45);
\draw (2.4,-1.25) -- (1.85,-1.6);
\draw (2.4,-1.75) -- (1.85,-1.80);
\draw (2.4,-1.75) -- (1.85,-1.95);
\draw (2.4,-1.75) -- (1.85,-2.1);
\draw (2.4,-3.5) -- (1.85,-3.05);
\draw (2.4,-3.5) -- (1.85,-3.2);
\draw (2.4,-3.5) -- (1.85,-3.35);

\draw (0.35,-2.5) -- (2.4,-2.75);
\draw[loosely dotted]  (0.35,-1.65) -- (0.35,-2.35);
\draw[loosely dotted]  (0.35,-2.55) -- (0.35,-3.45);
\draw[loosely dotted]  (2.4,-1.85) -- (2.4,-2.65);
\draw[loosely dotted]  (2.4,-2.85) -- (2.4,-3.45);
\end{tikzpicture}

\caption{Predictive correlation screening thresholds  the matrix $\bH^{xy}$ in \eqref{eq:CoefMat} to find variables $X_i$ that are most predictive of responses  $Y_j$. This is equivalent to finding sparsity in a bipartite graph ${\mathcal G}_{\rho^*}({\mathbf \bH^{xy}})$ with parts $x$ and $y$ which have $p$ and $q$ vertices, respectively. For $1 \leq i \leq p$ and $1 \leq j \leq q$, vertex $X_i$ in part $x$ is connected to vertex $Y_j$ in part $y$ if
$|h_{ij}^{xy}| > \rho^*$.} \label{fig:bipartite_graph}
\end{center}
\end{figure}

Under the assumptions of Propositions \ref{Prop1} and \ref{Prop2}:
\be p(N_{\delta, \rho_p}^{xy}>0) \rightarrow
1-\exp(-\xi_{p,q,n,\delta,\rho_p}) ~~\text{as} ~~ p \rightarrow
\infty
 \ee
Using the above limit, approximate p-values can be computed.
Fix a degree threshold $\delta \leq q$ and a correlation threshold
$\rho^* \in [0,1]$. Define ${\mathcal G}_{\rho^*}({\mathbf
\bH^{xy}})$ as the undirected bipartite graph (Fig. \ref{fig:bipartite_graph}) with parts labeled
$x$ and $y$, vertices $\{X_1,X_2,...,X_p \}$ in part $x$ and
$\{Y_1,Y_2,...,Y_q \}$ in part $y$. For $1 \leq i \leq p$ and $1
\leq j \leq q$, vertices $X_i$ and
$Y_j$ are connected if
$|h_{ij}^{xy}|
>\rho^*$, where $h_{ij}^{xy}$ is the $(i,j)$th entry of $\bH^{xy}$ defined in
\eqref{Hxy}. Denote by $d_i^x$ the degree of vertex $X_i$ in ${\mathcal
G}_{\rho^*}({\mathbf \bH^{xy}})$. For each
value $\delta \in \{1,\cdots,\max_{1 \leq i \leq p} d^x_i \}$, and each $i$, $1 \leq i \leq p$, denote by
$\rho_i(\delta)$ the maximum value of the correlation threshold $\rho$
for which $d^x_i \geq \delta$ in ${\mathcal G}_\rho({\mathbf
\bH^{xy}})$. $\rho_i(\delta)$ is in fact equal to the $\delta$th
largest value $|h_{ij}^{xy}|, 1 \leq j \leq q$. $\rho_i(\delta)$ can be computed using
Approximate Nearest Neighbors (ANN) type algorithms
\citep{jegou2011product,arya1998optimal}. Now for each $i$ define the
modified threshold $\rho_i^{\text{mod}}(\delta)$ as: \be
\rho_i^{\text{mod}}(\delta) = w_i \rho_i(\delta) , \;\;\;\;\;1\leq i\leq p, \label{rhomod}
\ee where $w_i = D(i)/\sum_{j=1}^{p} D(j)$, in which $D(i)$ is
the $i$th diagonal element of the diagonal matrix
$\bD_{(\mU^x)^T(\mU^x(\mU^x)^T)^{-2}\mU^x}^{\frac{1}{2}}$ (recall Sec. \ref{sec:Prelim}).

Using Propositions 1 and 2 the p-value associated with variable $X_i$ at
degree level $\delta$ can be approximated as: \be pv_{\delta}(i) \approx
1-\exp(-\xi_{p,q,n,\delta,\rho_i^{\text{mod}}(\delta)}). \label{eq:p-vals} \ee The
set of p-values \eqref{eq:p-vals}, $i=1,\ldots, p$,  provides a measure of importance of
each variable $X_i$ in predicting $Y_j$'s. Under a block-sparsity null hypothesis, the most
important variables would be the ones that have the smallest
p-values.
Similar to the result in \citep{hero2011large,hero2012hub}, there is a
phase transition in the p-values as a function of threshold $\rho$. More
exactly, there is a critical threshold $\rho_{c,\delta}$ such
that if $\rho > \rho_{c,\delta}$, the average
number $E[N_{\delta,\rho}^{xy}]$ of discoveries abruptly decreases to $0$ and if $\rho < \rho_{c,\delta}$ the average
number of discoveries abruptly increases to $p$. The value of this critical threshold is:  \be
\rho_{c,\delta}=\sqrt{1-(c_{n,\delta}^{xy}p)^{-2\delta/(\delta(n-2)-2)}},
\label{critical} \ee where $c_{n,\delta}^{xy}=\ra_n \delta
J(\ol{f_{\bU_{\bullet}^x,\bU_{\ast_1}^y,
\ldots,\bU_{\ast_{\delta}}^y}})$. When $\delta=1$, the expression given in \eqref{critical} is identical, except for the constant $c_{n,\delta}^{xy}$, to the expression $(3.14)$ in \citep{hero2011large}. 

Expression (\ref{critical}) is useful in choosing the PCS correlation threshold
$\rho^*$. Selecting $\rho^*$ slightly greater than
$\rho_{c,\delta}$ will prevent the bipartite
graph ${\mathcal G}_{\rho^*}({\mathbf \bH^{xy}})$ from having an overwhelming number of edges.

Normally $\delta=1$ would be selected to find all regressor variables predictive of at least 1 response variable $Y_j$. A value of $\delta=d>1$ would be used if the experimenter were only interested in variables that were predictive of at least $d$ of the responses. Pseudo-code for the complete algorithm for variable selection is shown in Fig. \ref{fig:PCS}. The worse case computational complexity of the PCS algorithm is only $O(n p \log q)$. \\

\begin{figure}[!h]
\begin{center}


\vrule
\begin{minipage}{8.5cm} 
\hrule \vspace{0.5em} 
\begin{minipage}{7.5cm} 
        \begin{itemize}
        \item Initialization:

           \begin{enumerate}

             \item Choose an initial threshold $\rho^*>\rho_{c,\delta}$

             \item Calculate the degree of each vertex on side $x$ of the bipartite graph ${\mathcal G}_{\rho^*}({\mathbf \bH^{xy}})$
			
			\item Select a value of $\delta \in \{1,\cdots,\max_{1 \leq i \leq p} d^x_i\}$
			
           \end{enumerate}

        	\item For each $i = 1,\cdots, p$ find $\rho_i(\delta)$ as the $\delta$th greatest element of $\{|h_{ij}|, 1 \leq j \leq q \}$
        	
        	\item Compute $\rho_i^{\text{mod}}(\delta)$ using (\ref{rhomod})
        	
        	\item Approximate the p-value corresponding to the $i$th independent
variable $X_i$ as $pv_{\delta}(i) \approx 1-\exp(-\xi_{p,q,n,\delta,\rho_i^{\text{mod}}(\delta)})$.

\item Screen variables by thresholding the p-values $pv_\delta(i)$ at desired significance level

        \end{itemize}

\end{minipage}
\vspace{0.5em} \hrule
\end{minipage}\vrule \\
\end{center}
\caption{Predicive Correlation Screening (PCS) Algorithm} \label{fig:PCS}
\end{figure}

\section{Two-stage predictor design}
\label{sec:two-stage}
Assume there are a total of $t$ samples $\{\bfY_i,\bfX_i\}_{i=1}^t$ available. During the first stage a number $n\leq t$ of these samples are assayed for all $p$ variables and during the second stage the rest of the $t-n$ samples are assayed for a subset of $k\leq p$ of the variables. Subsequently, a $k$-variable predictor is designed using all $t$ samples collected during both stages. The first stage of the PCS predictor is implemented by using the PCS algorithm with $\delta=1$.

As this two-stage PCS algorithm uses $n$ and $t$ samples in stage $1$ and stage $2$ respectively, we denote the algorithm above as the $n|t$ algorithm. Experimental results in Sec. \ref{sec:Simulations}  show that for $n \ll p$, if LASSO or correlation learning is used instead of PCS in stage 1 of the two-stage predictor the performance suffers. An asymptotic analysis (as the total number of samples $t \rightarrow \infty$) of the above two-stage predictor can be performed to obtain optimal sample allocation rules for stage 1 and stage 2. The asymptotic analysis discussed in Sec. \ref{sec:Asymptotic} provides minimum Mean Squared Error (MSE) under the assumption that $n$, $t$, $p$, and $k$ satisfy the budget constraint:
\begin{equation}
np + (t-n)k \leq \mu,
\label{eq:assaycostcond}
\end{equation}
where $\mu$ is the total budget available. The motivation for this condition is to bound the total sampling cost of the experiment.

\section{Optimal stage-wise sample allocation}
\label{sec:Asymptotic}
We first give theoretical upper bounds on the Family-Wise Error Rate (FWER) of performing variable selection using p-values obtained via PCS. Then, using the obtained bound, we compute the asymptotic optimal sample size $n$ used in the first stage of the two-stage predictor, introduced in the previous section, to minimize the asymptotic expected MSE. 

We assume that the response $\bY$ satisfies the following ground truth model: \be \bY = \ba_{i_1} X_{i_1} + \ba_{i_2} X_{i_2} + \cdots + \ba_{i_k} X_{i_k} + \bN , \ee where $\pi_0 = \{ i_1,\cdots, i_k \}$ is a set of distinct indices in $\{1,\ldots,p \}$, $\bX = [X_1, X_2, \cdots, X_p]$ is the vector of predictors, $\bY$ is the
$q$-dimensional response vector, and $\bN$ is a noise vector statistically independent of $\bX$. $X_{i_1},\cdots,X_{i_k}$ are called active variables and the remaining $p-k$ variables are called inactive variables. We assume that the $p$-dimensional vector $\bX$ follows a multivariate normal distribution with mean $\mathbf{0}$ and $p \times p$ covariance matrix $\mathbf{\Sigma} = [\sigma_{ij}]_{1 \leq i,j \leq p}$, where $\mathbf{\Sigma}$ has the following block diagonal structure:
\be
\sigma_{ij} = \sigma_{ji} = 0, ~~~\forall ~ i \in \pi_0, j \in \{1,\cdots,p \} \backslash \pi_0.
\label{SigmaProperty}
\ee
In other words active (respectively inactive) variables are only correlated with the other active (respectively inactive) variables. Also, we assume that $\bN$ follows a multivariate normal distribution with mean $\mathbf{0}$ and covariance matrix $\sigma \bI_{q \times q}$.

We use the PCS algorithm of Sec. \ref{sec:Predictive} with $\delta=1$ to select the $k$ variables with the smallest p-values. These selected variables will then be used as estimated active variables in the second stage. The following proposition gives an upper bound on the probability of selection error for the PCS algorithm.

\begin{propositions}
\label{Prop:UpperBound}
If $n \geq \Theta(\log p)$ then with probability at least $1-q/p$, PCS recovers the exact support $\pi_0$.
\end{propositions}
\noindent{\it Proof of Proposition \ref{Prop:UpperBound}:} See appendix. \qed

Proposition 3 can be compared to Thm. $1$ in \citep{obozinski2008high} for recovering the support $\pi_0$ by minimizing a LASSO-type objective function. 
The constant in $\Theta(\log p)$ of Prop. 3 is increasing in the dynamic range coefficient
\be
\max_{i=1,\cdots,q} \frac{|\pi_0|^{-1}\sum_{j \in \pi_0} |b_{ij}|}{\min_{j \in \pi_0} |b_{ij}|} \in [1, \infinity) \label{quantity1},
\label{drc}
\ee
where $\mathbf B = [\mathbf{b}_1, \cdots, \mathbf{b}_p] = \mathbf{\Sigma}^{1/2} \bA$. The worst case (largest constant in $\Theta(\log p)$) occurs when there is high dynamic range in some rows of the $q \times p$ matrix $\bB$. 

The following proposition states the optimal sample allocation rule for the two-stage predictor, as $t \rightarrow \infty$.

\begin{propositions}
\label{Prop:UpperBound2}
The optimal sample allocation rule for the two-stage predictor introduced in Sec. \ref{sec:two-stage} under the cost condition \eqref{eq:assaycostcond} is 
\be n = \left\{\begin{array}{cc} O(\log t), & c(p-k) \log t+k t \leq \mu\\
0, & o.w.
\end{array} \right.
\ee
\end{propositions}
\noindent{\it Proof of Proposition \ref{Prop:UpperBound2}:} See appendix. \qed

\begin{figure}[p]
\centering
\subfigure {\includegraphics[width=3in]{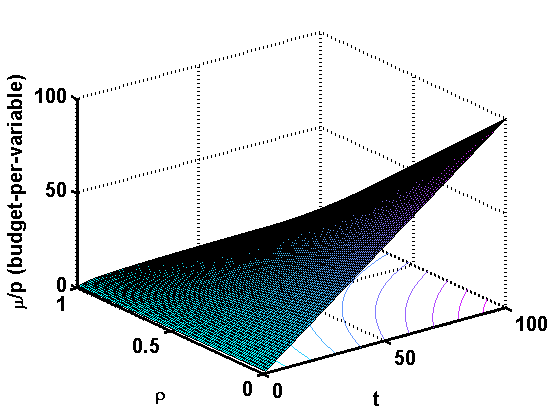}}
\subfigure {\includegraphics[width=3in]{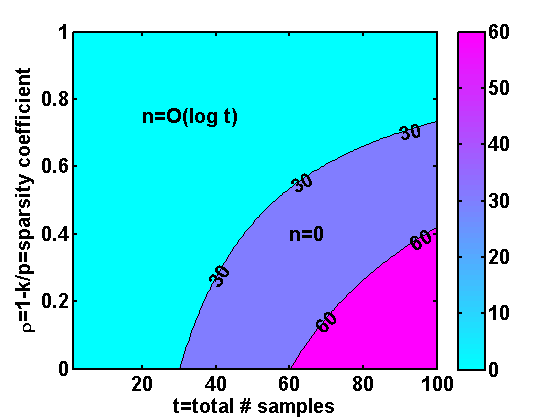}}
\caption{Left: surface $\mu/p=\rho \log t+(1-\rho) t$. Right: contours indicating optimal allocation regions for $\mu/p=30$ and $\mu/p=60$. ($\rho=1-k/p$)}
\label{fig:budget}
\end{figure}

Proposition 4 implies that for a generous budget ($\mu$ large) the optimal first stage sampling allocation is $\log(t)$. However, when the budget is tight it is better to skip stage 1 ($n=0$). Figure \ref{fig:budget} illustrates the allocation region as a function of the sparsity coefficient $\rho=1-k/p$.

\section{Simulation results}
\label{sec:Simulations}
\paragraph{Efficiency of Predictive Correlation Screening.}

We illustrate the performance of the two-stage PCS algorithm and compare to LASSO and correlation learning methods \citep{tibshirani1996regression,genovese2012comparison}. 

In the first set of simulations we generated an $n \times
p$ data matrix $\mX$ with independent columns, each of which is
drawn from a $p$-dimensional multivariate normal distribution with
identity covariance matrix. The $q \times p$ coefficient matrix
$\bA$ is then generated in such a way that each column of $\bA$ is
active with probability $0.1$. Each active column of $\bA$ is a
random $q$-dimensional vector with i.i.d. $\mathcal{N}(0,1)$
entries, and each inactive column of $\bA$ is a zero vector.
Finally, a synthetic response matrix $\mY$ is generated
by a simple linear model \be \mY^T = \bA \mX^T + \mN^T, \label{lm} \ee 
where $\mN$ is $n \times q$ noise matrix whose entries are i.i.d.
$\mathcal{N}(0,0.05)$. The importance of a variable is measured
by the value of the $\ell_2$ norm of the corresponding column of $\bA$. Note that
the linear model in \eqref{lm} trivially satisfies the block sparsity
assumptions on the covariance matrices in Prop. \ref{Prop2}.

\begin{figure} 
\centering
\includegraphics[width=3.0in]{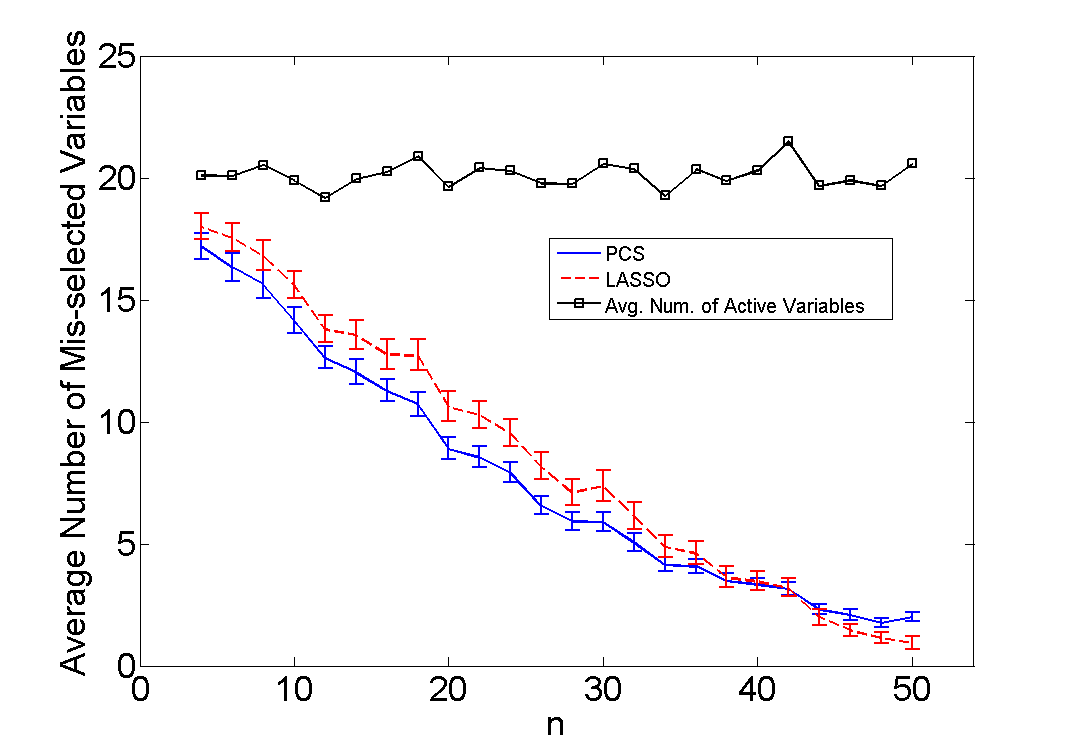}
\caption{Average number of mis-selected variables for active set
implementation of LASSO (dashed) vs. Predictive Correlation
Screening (solid), $p=200, q=20$.} \label{fig:SALSA_Vars}
\end{figure}

\begin{figure} 
\centering
\includegraphics[width=3.0in]{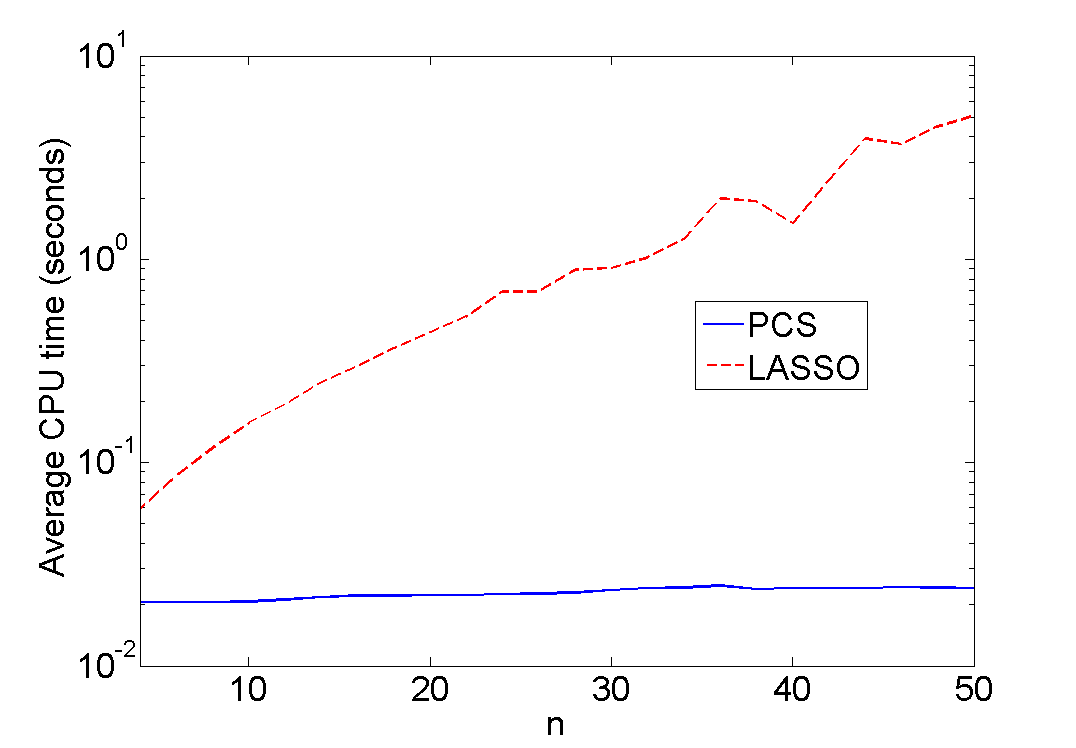}
\caption{Average CPU time for active set implementation of LASSO (dashed)
vs. PCS (solid), $p=200, q=20$.}
\label{fig:SALSA_Time}
\end{figure}

We implemented LASSO using an active set type algorithm - claimed to be one the fastest methods for solving LASSO \citep{kim2010fast}. We set the number of regressor and response variables to $p=200$ and $q=20$, respectively, while the number
of samples $n$ was varied from $4$ to $50$. Figure
\ref{fig:SALSA_Vars} shows the average number of mis-selected variables
for both methods, as a function of $n$. The plot is computed by
averaging the results of $400$ independent experiments for each
value of $n$. Figure \ref{fig:SALSA_Time} shows
the average run time on a logarithmic scale, as a function of $n$ (MATLAB version 7.14 running on 2.80GHz CPU). As we see, for low number of samples, PCS has  better performance than LASSO and is significantly faster.

\begin{figure} 
\centering
\includegraphics[width=3.0in]{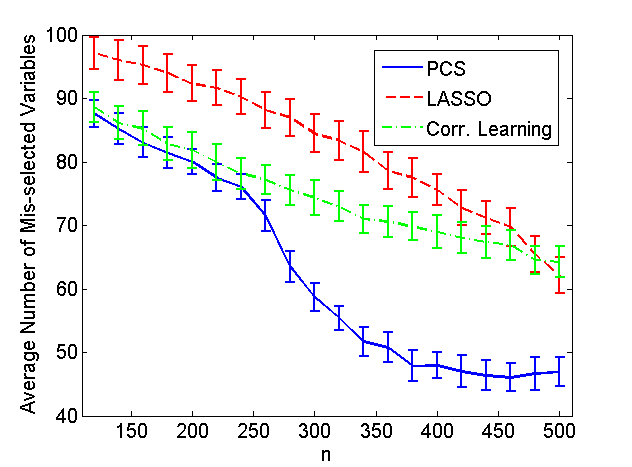}
\caption{Average number of mis-selected variables. Active set implementation of LASSO (red-dashed)
vs. correlation learning (green-dashed) vs. PCS (solid), $p=10^4, q=1$.}
\label{fig:LASSO100}
\end{figure}

To illustrate PCS for a higher dimensional example, we set $p=10^4, q=1$ and
compared PCS with LASSO and also the correlation learning method of \citep{genovese2012comparison}, for a small number of samples. Figure
\ref{fig:LASSO100} shows the results of this simulation over
an average of $400$ independent experiments for each value of $n$. In this experiment, exactly $100$ entries of $\bA$ are active. The active entries are i.i.d. draws of $\mathcal{N}(0,1)$ and inactive entries are equal to zero. Unlike Fig.  \ref{fig:SALSA_Vars}, here the regressors variables are correlated. Specifically, $X_1,\cdots,X_p$, are i.i.d. draws from a multivariate normal distribution with mean $\mathbf{0}$ and block diagonal covariance matrix satisfying \eqref{SigmaProperty}. As we see for small number of samples, PCS performs significantly better in selecting the important regressor variables. 

\paragraph{Efficiency of The Two-stage Predictor.}

\begin{figure} [ht]
\centering
\includegraphics[width=3.0in]{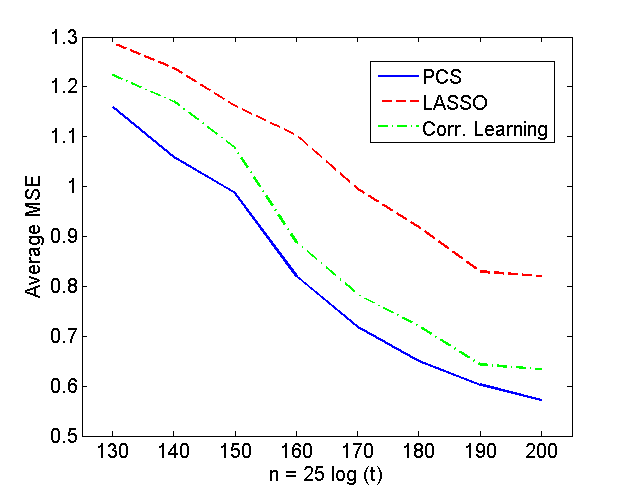}
\caption{Prediction MSE for the two-stage predictor when $n = 25 \log t$ samples are used  for screening at the first stage and all $t$ samples are used for computing the OLS estimator coefficients at the second stage. The solid plot shows the MSE when PCS is used in the first stage while the red and green dashed plots show the MSE when PCS is replaced with LASSO and correlation learning, respectively. Here, $p=10^4$ and $q=1$. The Oracle OLS (not shown), which is the OLS predictor constructed on the true support set, has average MSE performance that is a factor of 2 lower than the curves shown in the figure. This is due to the relatively small sample size available to these algorithms.}
\label{fig:p100002n_MSE}
\end{figure}

To test the efficiency of the proposed two-stage predictor, a total of $t$ samples are generated using the linear model \eqref{lm} from which $n = 25 \log t$ are used for the task of variable selection at the first stage. All $t$ samples are then used to
compute the OLS estimator restricted to the selected variables. We chose $t$ such that $n = (130:10:200)$. The performance is evaluated by the empirical MSE:= $\sum_{i=1}^{m} (Y_i-\hat{Y}_i)^2 / m,$
where $m$ is the number of simulation trials. Similar to the previous experiment, exactly $100$ entries of $\bA$ are active and the regressor variables follow a multivariate normal distribution with mean $\mathbf{0}$ and block diagonal covariance matrix of the form \eqref{SigmaProperty}. 
Figure \ref{fig:p100002n_MSE} shows
the result of this simulation for $p=10^4$ and $q =1$.
Each point on the plot is an average of $100$ independent
experiments. Observe that in this low sample regime, when LASSO or correlation learning are used instead of PCS in the first stage, the performance suffers.
\paragraph{Estimation of FWER Using Monte Carlo Simulation.}

\begin{figure} [h]
\centering
\includegraphics[width=3.0in]{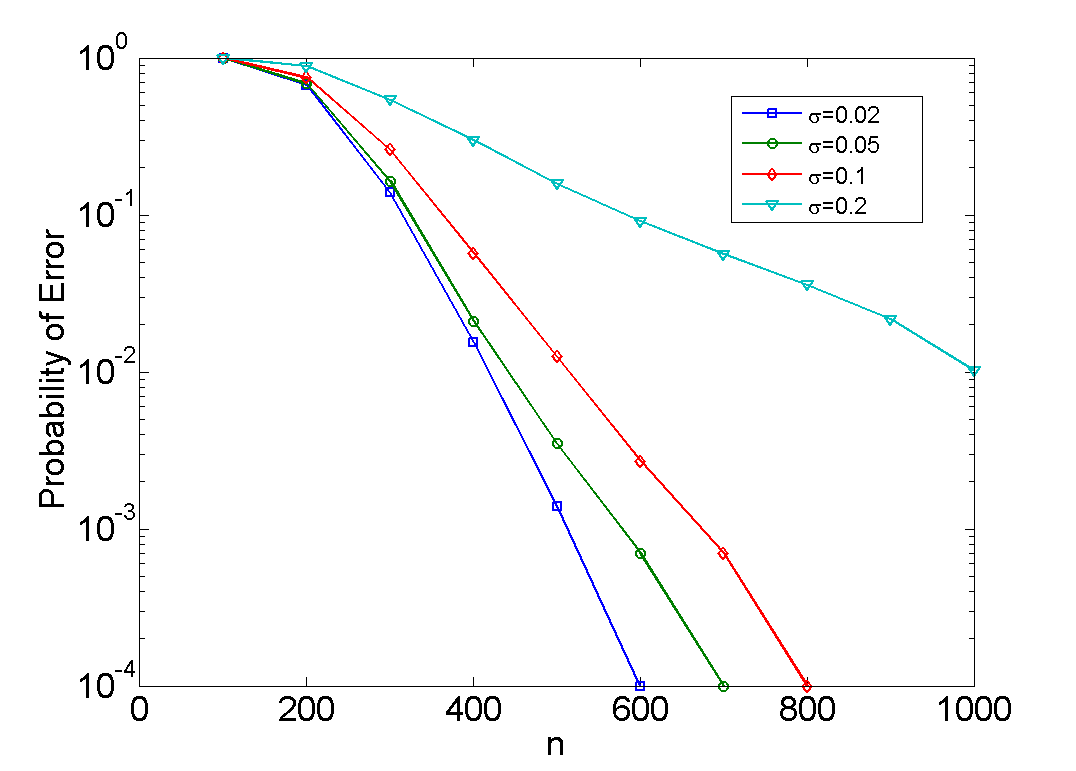}
\caption{Probability of selection error as a function of number of samples for PCS. The entries of the coefficient matrix are i.i.d. draws from distribution \eqref{BerGauss}.}
\label{fig:UpperBound_logy}
\end{figure}

We set $p = 1000, k=10$ and $n=(100:100:1000)$ and using Monte Carlo simulation, we computed the probability of error (i.e. when the exact support is not recovered) for the PCS. In order to prevent the ratios $|a_{j}|/\sum_{l \in \pi_0} |a_{l}|, j \in \pi_0$ from getting close to zero, the active coefficients were generated via a Bernoulli-Gaussian distribution of the form:
\begin{equation}
a \sim 0.5 \mathcal{N}(1,\sigma^2) + 0.5 \mathcal{N}(-1,\sigma^2),
\label{BerGauss}
\end{equation}
Figure \ref{fig:UpperBound_logy} shows the estimated probabilities. Each point of the plot is an average of $N=10^4$ experiments. As the value of $\sigma$ decreases dynamic range coefficient \eqref{drc} goes to infinity with high probability and the probability of selection error degrades. As we can see, the FWER decreases at least exponentially with the number of samples. This behavior is consistent with Prop. \ref{Prop:UpperBound}.
\paragraph{Application to Experimental Data.}

We illustrate the application of the proposed two-stage predictor on the  Predictive Health and Disease dataset, which consists of gene expression levels and symptom scores of $38$ different subjects. The data was collected during a challenge study for which  some subjects become symptomatically ill with the H3N2 flu virus \citep{huang2011temporal}. For each subject, the gene expression levels and the symptoms have been recorded at a large number of time points that include pre-inoculation and post-inoculation sample times. $10$ different symptom scores were measured. Each symptom score takes an integer value from $0$ to $4$, which measures the severity of that symptom at the corresponding time. 
The goal here is to learn a predictor that can accurately predict the symptom scores of a subject based on his measured gene expression levels.

The number of predictor variables (genes) selected in the first stage is restricted to $50$. Since, the symptom scores take integer values, the second stage uses multinomial logistic regression instead of the OLS predictor. The performance is evaluated by leave-one-out cross validation. To do this, the data from all except one subject are used as training samples and the data from the remaining subject are used as the test samples. The final MSE is then computed as the average over the $38$ different leave-one-out cross validation trials. In each of the experiments $18$ out of the $37$ subjects of the training set, are used in first stage and all of the $37$ subjects are used in the second stage. It is notable that except for the first two symptoms, PCS performs better in predicting the symptom scores.

Note that, in this experiment, each symptom is considered as a one dimensional response and the two-stage algorithm is applied to each symptom separately. 
\begin{table}[h] 
\center
\begin{tabular}{|l|l|l|}
  \hline
  Symptom & MSE: LASSO & MSE: PCS\\
  \hline
  Runny Nose & 0.3346 & 0.3537\\
  Stuffy Nose & 0.5145 & 0.5812\\
  Sneezing & 0.4946 & 0.3662\\
  Sore Throat & 0.3602 & 0.3026\\
  Earache & 0.0890 & 0.0761\\
  Malaise & 0.4840 & 0.3977\\
  Cough & 0.2793 & 0.2150\\
  Shortness of Breath & 0.1630 & 0.1074\\
  Headache & 0.3966 & 0.3299\\
  Myalgia & 0.3663 & 0.3060\\
  \hline
  Average for all symptoms & 0.3482 & 0.3036\\
  \hline
\end{tabular}
\caption{MSE of the two-stage LASSO predictor and the proposed two-stage PCS predictor used for symptom score prediction. The data come from a challenge study experiment that collected gene expression and symptom data from human subjects \citep{huang2011temporal}.}
\label{table:PHD}
\end{table}

\section{Conclusion}
We proposed an algorithm called Predictive Correlation Screening (PCS) for approximating the p-values of candidate predictor variables in high dimensional linear regression under a sparse null hypothesis. Variable selection was then performed based on the approximated p-values. PCS is specifically useful in cases where $n \ll p$  and the high cost of assaying all regressor variables justifies a two-stage design: high throughput variable selection followed by predictor construction using fewer selected variables. Asymptotic analysis and experiments showed advantages of PCS as compared to LASSO and correlation learning.

\clearpage

\bibliography{Refs}
\bibliographystyle{icml2013}

\clearpage

\section{Appendix}

\noindent{\it Proof of Prop. \ref{Prop1}:}

Define $\phi_i^x = I(d_i^x \geq \delta)$, where $d_i^x$ is the
degree of vertex $i$ in part $x$ in the thresholded
correlation graph. We have: $N_{\delta, \rho}^{xy} =
\sum_{i=1}^{p} \phi_i^x$. Define $\phi_{ij}^{xy}=I(\bU_j^y \in
A(r,\bU_i^x))$, where $A(r,\bU_i^x)$ is the union of two anti-polar
caps in $S_{n-2}$ of radius $\sqrt{2(1-\rho)}$ centered at $\bU_i^x$
and $-\bU_i^x$. $\phi_i^x$ can be expressed as: \be \phi_i^x=
\sum_{l=\delta}^{q} \sum_{\vec{k} \in \breve{C}(q,l)}
\prod_{j=1}^{l} \phi_{ik_j}^{xy} \prod_{m=l+1}^{q} (1-\phi_{ik_m}
^{xy}), \ee where $\vec{k}=(k_1,...,k_q)$ and $\breve{C}(q,l)= \{
\vec{k}: k_1 < k_2 < ... <k_l, k_{l+1}< ... < k_q, k_j \in
\{1,2,...,q\}, k_i
\neq k_j \}$.\\
By subtracting $\sum_{\vec{k} \in \breve{C}(q,l)}
\prod_{j=1}^{\delta} \phi_{ik_j}^{xy}$ from both sides, we get:
\begin{eqnarray} \label{phidiff}
\phi_i^x - \sum_{ \vec{k} \in \breve{C}(q,l)}
\prod_{j=1}^{\delta} \phi_{ik_j}^{xy} =  \nonumber \\
\sum_{l=\delta+1}^{q} \sum_{\vec{k} \in \breve{C}(q,l)}
\prod_{j=1}^{l} \phi_{ik_j}^{xy} \prod_{m=l+1}^{q} (1-\phi_{ik_m}
^{xy}) + \nonumber \\ \sum_{\vec{k} \in \breve{C}(q,\delta)}
\sum_{m=\delta+1}^{q} (-1)^{m-\delta}  \prod_{j=1}^{\delta}
\phi_{ik_j}^{xy} \nonumber \\ \sum_{k'_{\delta+1}<...<k'_m, \{
k'_{\delta+1},...,k'_m \} \subset \{ k_{\delta+1},...,k_q \} }
\prod_{n=\delta +1}^{m} \phi_{ik'_n}^{xy}.
\end{eqnarray}
The following inequality will be helpful: \begin{eqnarray}
\label{exprep2} E[\prod_{i=1}^{k} \phi_{ii_j}^{xy}] =
\int_{S_{n-2}} dv \int_{A(r,v)} du_1 ... \int_{A(r,v)}du_k \nonumber \\
f_{U_{i_1}^y,...,U_{i_k}^y,U_i^x} (u_1,...,u_k,v) \end{eqnarray}
\be \label{expbound2} \leq P_0^k a_n^k M_{K|1}^{yx}, \ee where
$M_{K|1}^{yx} = \text{max}_{i_1 \neq ... \neq i_k, i} \|
f_{U_{i_1}^y,...,U_{i_k}^y|U_i^x} \|_{\infty}$.\\
Also we have: \be \label{Qbound} E[\prod_{l=1}^{m}
\phi_{i_lj_l}^{xy}] \leq P_0^m a_n^m M_{|Q|}^{yx}, \ee where
$Q=unique(\{i_l, j_l\})$ is the set of unique indices among the
distinct pairs $\{ \{i_l, j_l\} \}_{l=1}^{m}$ and $M_{|Q|}^{yx}$
is
a bound on the joint density of $\bU_Q^{xy}$.\\
Now define: \be \theta_i^x=\binom{q}{\delta}^{-1} \sum_{\vec{k}
\in \breve{C}(q,\delta)} \prod_{j=1}^{\delta} \phi_{ik_j}^{xy}. \ee
Now, we show that \be |E[\phi_i^x] -
\binom{q}{\delta}E[\theta_i^x]| \leq
\gamma_{q,\delta}(qP_0)^{\delta+1}, \ee where $\gamma_{q,\delta} =
2 e \max_{\delta+1 \leq l \leq q} \{ a_n^l M_{l|1}^{yx} \}$. To
show this, take expectations from both sides of equation
\eqref{phidiff} and apply the bound \eqref{expbound2} to obtain:
\begin{eqnarray} \label{diffbound}
|E[\phi_i^x-\binom{q}{\delta} E[\theta_i^x]]|  \nonumber \\ \leq
\sum_{l=\delta+1}^q \binom{q}{l}P_0^la_n^lM_{l|1}^{yx} + \nonumber \\
\nonumber \binom{q}{\delta} \sum_{l=1}^{q-\delta}
\binom{q-\delta}{l}P_0^{\delta+l}a_n^{\delta+l}M_{\delta+l|1}^{yx}
\nonumber \\ \nonumber \leq \text{max}_{\delta+1 \leq l \leq q}
\{a_n^l M_{l|1}^{yx} \} \\ \nonumber (\sum_{l=\delta+1}^{q}
\binom{q}{l} P_0^{l} + \binom{q}{\delta}
P_0^{\delta} \sum_{l=1}^{q-\delta} \binom{q-\delta}{l} P_0^l) \\
\nonumber \leq \text{max}_{\delta+1 \leq l \leq q} \{a_n^l
M_{l|1}^{yx} \} \\ \nonumber ((e-\sum_{l=1}^{\delta}\frac{1}{l!})
(qP_0)^{\delta+1} + \frac{q^{\delta}}{\delta!} P_0^\delta (e-1)
(q-\delta)P_0) \\ \leq  \text{max}_{\delta+1 \leq l \leq q}
\{a_n^l M_{l|1}^{yx} \} 2e (qP_0)^{\delta+1},
\end{eqnarray}
in which, the third inequality follows from the assumption $q P_0
\leq 1$ along with the inequality : \be \label{suminq}
\sum_{k=s+1}^{G} \binom{G}{k}(\frac{t}{G})^k \leq \sum_{k=s+1}^G
\frac{t^k}{k!} \\ \nonumber \leq (e-\sum_{k=0}^{s}\frac{1}{k!})t^{s+1},~ 0 \leq
t \leq 1. \ee Application of the mean value theorem to the integral
representation \eqref{exprep2} yields: \be \label{thetadiffbound}
|E[\theta_i^x] - P_0^{\delta}
J(\overline{f_{\bU_{*_1}^y,...,\bU_{*_\delta}^y,\bU_i^x }})| \leq
\tilde{\gamma}^{yx}_{q,\delta}(qP_0)^{\delta}r, \ee where
$\tilde{\gamma}^{yx}_{q,\delta}=2a_n^{\delta+1}\dot{M}^{yx}_{\delta+1|1}/\delta!$
and $\dot{M}^{yx}_{\delta+1|1}$ is a bound on the norm of the
gradient: \be \nabla_{\bU_{i_1}^y,...,\bU_{i_{\delta}}^y}
\overline{f_{\bU_{*_1}^y,...,\bU_{*_\delta}^y|\bU_i^x
}}(\bU_{i_1}^y,...,\bU_{i_{\delta}}^y|\bU_i^x). \ee Combining
\eqref{diffbound} and \eqref{thetadiffbound} and using the
relation $r=O((1-\rho)^{1/2})$ we conclude: \begin{eqnarray}
|E[\phi_i^x]-\binom{q}{\delta}P_0^{\delta}
J(\overline{f_{\bU_i^x,\bU_{*_1}^y,...,\bU_{*_\delta}^y }})| \leq
\nonumber \\ O(p^{\delta}(qP_0)^{\delta} \text{max} \{
pP_0,(1-\rho)^{1/2}\}).
\end{eqnarray} Summing up over $i$ we conclude:
\begin{eqnarray}
E[N_{\delta,\rho}^{xy}]-\xi_{p,q,n,\delta,\rho}^{xy}
J(\overline{f_{\bU_{*}^x,\bU_{\bullet 1}^y,...,\bU_{\bullet \delta}^y}})
\leq \nonumber \\  O(p(pP_0)^{\delta} \text{max} \{ pP_0,(1-\rho)^{1/2}) \\
\nonumber = O( (\eta_{p,q,\delta}^{xy})^{\delta} \text{max} \{
\eta_{p,q,\delta}^{xy}p^{-\frac{1}{\delta}},(1-\rho)^{1/2} \}),
\end{eqnarray} where $\eta_{p,q,\delta}^{xy} = p^{1/\delta} q P_0$.
This concludes \eqref{prop1first}.\\
To prove the second part of the theorem, we use Chen-Stein method
\citep{arratia1990poisson}. Define: \be \tilde{N}^{xy}_{\delta, \rho} =
\sum_{0 \leq i_0 \leq p, 0 \leq i_1 < ... <i_{\delta} \leq q}
\prod_{j=1}^{\delta} \phi_{i_0i_j}^{xy}. \ee Assume the vertices
$i$ in part $x$ and $y$ of the thresholded graph are shown by
$i^x$ and $i^y$ respectively. for
$\vec{i}=(i_0^x,i_1^y,...,i_{\delta}^y)$, define the index set
$B^{xy}_{\vec{i}} = B^{xy}_{(i_0^x,i_1^y,...,i_{\delta}^y)} = \{
(j_0^x,j_1^y...,j_{\delta}^y): j_1^x \in
\mathcal{N}_{k}^{xy}(i_1^x) \cup i_1^x, j_l^y \in
\mathcal{N}_{k}^{xy}(i_l^y) \cup i_l^y, l=1,...,\delta \} \cap
C^{xy}_{<}$ where $C^{xy}_{<} = \{(j_0,...,j_{\delta}): 1 \leq j_0
\leq p,1 \leq j_1 <...<j_{\delta} \leq q \}$. Note that
$|B_{\vec{i}}^{xy}| \leq k^{\delta+1}$. We have: \be
\label{Ntilde} \tilde{N}^{xy}_{\delta, \rho} = \sum_{\vec{i} \in
C^{xy}_{<}} \prod_{j=1}^{\delta} \phi_{i_0i_j}^{xy}. \ee Assume
$N^{*xy}_{\delta, \rho}$ is a Poisson random variable with
$E[N^{*xy}_{\delta, \rho}] = \tilde{N}^{xy}_{\delta, \rho}$. Using
theorem 1 of \citep{arratia1990poisson}, we have: \be 2 ~\text{max}_A
|p(\tilde{N}^{xy}_{\delta, \rho} \in A ) -
p(\tilde{N}^{*xy}_{\delta, \rho} \in A)| \leq b_1 + b_2 + b_3, \ee
where: \be b_1 = \sum_{\vec{i} \in C^{xy}_{<}} \sum_{\vec{j} \in
B_{\vec{i}}^{xy} -\vec{i}}  E[\prod_{l=1}^{\delta}
\phi_{i_0i_l}^{xy}] E[ \prod_{m=1}^{\delta} \phi_{j_0j_m}^{xy}], \\
b_2 = \sum_{\vec{i} \in C^{xy}_{<}} \sum_{\vec{j} \in
B_{\vec{i}}^{xy} -\vec{i}}  E[\prod_{l=1}^{\delta}
\phi_{i_0i_l}^{xy} \prod_{m=1}^{\delta} \phi_{j_0j_m}^{xy}], \ee
and for $p_{\vec{i}^{xy}} = E[\prod_{l=1}^{\delta}
\phi_{i_0i_l}^{xy}]$: \be b_3 = \sum_{\vec{i} \in C^{xy}_{<}} E[
E[ \prod_{l=1}^{\delta} \phi_{i_0i_l}^{xy} - p_{\vec{i}^{xy}} |
\phi_{\vec{j}}^x : \vec{j} \not\in B_{\vec{i}}^{xy} ]]. \ee Using
the bound \eqref{Qbound}, $E[ \prod_{l=1}^{\delta}
\phi_{i_0i_l}^{xy}]$ is of order $O(P_0^{\delta})$. Therefore: \be
b_1 \leq O(pq^{\delta} k^{\delta +1} P_0^{2 \delta}) = \nonumber \\ =
O((\eta_{p,q,\delta}^{xy})^{2 \delta} (k/(p^{\frac{1}{\delta+1}}
q^{\frac{\delta}{\delta+1}}))^{\delta +1}). \ee Note that, since
$\vec{i} \neq \vec{j}, \prod_{l=1}^{\delta} \phi_{i_0i_l}^{xy}
\prod_{m=1}^{\delta} \phi_{j_0j_m}^{xy}$ is a multiplication of at
least $\delta +1$ different characteristic functions. Hence by
\eqref{Qbound}, \be E[\prod_{l=1}^{\delta} \phi_{i_0i_l}^{xy}
\prod_{m=1}^{\delta} \phi_{j_0j_m}^{xy}] = O(P_0^{\delta +1}). \ee
Hence, $b_2 \leq O(pq^{\delta} k^{\delta +1} P_0^{\delta +1}) =
O((\eta_{p,q,\delta}^{xy})^{\delta +1}
(k/(p^{\frac{1}{\delta}}q)^{1/(\delta+1)})^{\delta +1})$. Finally, to
bound $b_3$ we have:
\begin{eqnarray}
b_3 = \sum_{\vec{i} \in C_{<}^{xy}}  E[ E[ \prod_{l=1}^{\delta}
\phi_{i_0i_l}^{xy} - p_{\vec{i}^{xy}} | \bU_{A_{k}^{xy}(\vec{i})} ]]
= \\ \nonumber = \sum_{\vec{i} \in C_{<}^{xy}}
\int_{S_{n-2}^{|A_{k}^{xy}(\vec{i})|} dz_{A_{k}^{xy}(\vec{i})}}
(\prod_{l=1}^{\delta} \int_{S_{n-2}} dz_{i_0^x}
\int_{A(r,\bfu_{i_0}^x)} d\bfu_{i_l}^y) \\
\nonumber (\frac{f_{\bU_{\vec{i}}^{xy} |
\bU_{A_{k}^{xy}(\vec{i})}}(\bU_{\vec{i}}^{xy} |
\bU_{A_{k}^{xy}(\vec{i})}) -
f_{\bU_{\vec{i}}^{xy}}(\bU_{\vec{i}}^{xy})}{f_{\bU_{\vec{i}}^{xy}}(\bU_{\vec{i}}^{xy})}) \nonumber \\
f_{\bU_{\vec{i}}^{xy}}(\bU_{\vec{i}}^{xy})f_{\bU_{A_{k}^{xy}(\vec{i})}}(\bfu_{A_{k}^{xy}(\vec{i})})
\\ \leq \nonumber O(pq^{\delta} P_0^{\delta +1} \|
\Delta_{p,q,n,k,\delta}^{xy}\|_1)=
O((\eta_{p,q,\delta}^{xy})^{\delta} \|
\Delta_{p,q,n,k,\delta}^{xy}\|_1).
\end{eqnarray}
Therefore:
\begin{eqnarray} \label{Poissonbound}
|p(N_{\delta,\rho}^{xy} > 0)- (1-\text{exp}(-\Lambda^{xy}_{\delta}))| \leq \nonumber \\
|p(N_{\delta,\rho}^{xy} > 0)- (\tilde{N}_{\delta,\rho}^{xy} > 0)|
+ \nonumber \\ |p(\tilde{N}_{\delta,\rho}^{xy} > 0)-
(1-\text{exp}(-E[\tilde{N}_{\delta,\rho}^{xy}]))| + \nonumber \\
|\text{exp}(-E[\tilde{N}_{\delta,\rho}^{xy}]) -
\text{exp}(-\Lambda^{xy}_{\delta})| \nonumber \\ \leq 0 + b_1 + b_2 + b_3 +
O(|E[\tilde{N}_{\delta,\rho}^{xy}] -\Lambda^{xy}_{\delta}|).
\end{eqnarray}
Hence, it remains to bound $O(|E[\tilde{N}_{\delta,\rho}^{xy}]
-\Lambda^{xy}_{\delta}|)$. Application of mean value theorem to the multiple
integral \eqref{exprep2} gives: \be |E[\prod_{l=1}^{\delta}
\phi_{ii_l}^{xy}]-P_0^{\delta}J(f_{\bU_{i_1}^y,...,\bU_{i_{\delta}}^y,\bU_i^x})|
\leq O(P_0^{\delta}r). \ee Using relation \eqref{Ntilde} we
conclude: \begin{eqnarray} |E[\tilde{N}_{\delta,\rho}^{xy}] -
p\binom{q}{\delta}
P_0^{\delta}J(\overline{f_{\bU_{*_1}^y,...,\bU_{*_{\delta}}^y,\bU_{\bullet}^x}})|
\leq \nonumber \\ O(pq^{\delta}P_0^{\delta}r) =
O((\eta_{p,q,\delta}^{xy})^{\delta}r). \end{eqnarray} Combining
this with inequality \eqref{Poissonbound} along with the bounds on
$b_1, b_2$ and $b_3$, completes the proof of
\eqref{eq:Poissonconvcross}. \qed

\noindent{\it {Proof of Prop. \ref{Prop2}:}}

We prove the more general proposition below. Prop. \ref{Prop2} is then a direct consequence.

\noindent{\it {Proposition:}}
\noindent {\it Let $\mX$ and $\mY$ be $n\times p$ and $n \times q$ data matrices
whose rows are i.i.d. realizations of elliptically distributed
$p$-dimensional and $q$-dimensional vectors $\bX$ and $\bY$ with
mean parameters $\mathbf \mu_x$ and $\mathbf \mu_y$ and covariance
parameters $\mathbf \Sigma_{x}$ and $\mathbf \Sigma_{y}$,
respectively and cross covariance $\mathbf \Sigma_{xy}$. Let
$\mU^x=[\bU_1^x,\ldots, \bU_p^x]$ and $\mU^y=[\bU_1^y,\ldots,
\bU_q^y]$ be the matrices of correlation U-scores. Assume that the
covariance matrices $\mathbf \Sigma_{x}$ and $\mathbf \Sigma_{y}$
are block-sparse of degrees $d_x$ and $d_y$, respectively (i.e. by
rearranging their rows and columns, all non-diagonal entries are
zero except a $d_x \times d_x$ or a $d_y \times d_y$ block).
Assume also that the cross covariance matrix $\mathbf \Sigma^{xy}$
is block-sparse of degree $d_1$ for $x$ and degree $d_2$ for $y$
(i.e. by rearranging its rows and columns, all entries are zero
except a $d_1 \times d_2$ block), then
%
\be \tilde{\mU}^x=\mU^x(1+O(d_x/p)) . \label{tildeZ}\ee
Also assume that for $\delta\geq 1$ the joint density of any
distinct set of U-scores $\bU_{i}^x,\bU_{i_1}^y, \ldots,
\bU_{i_{\delta}}^y$ is bounded and differentiable over
$S_{n-2}^{\delta+1}$. Then the $(\delta+1)$-fold average function
$J(\ol{f_{\bU_{\bullet}^x,\bU_{\ast_1}^y, \ldots,
\bU_{\ast_{\delta}}^y}})$ and the average dependency coefficient
$\|\Delta_{p,n,k,\delta}^{xy}\|$ satisfy \be
J(\ol{f_{\bU_{\bullet}^x,\bU_{\ast_1}^y, \ldots,
\bU_{\ast_{\delta}}^y}})=1 +O(\max
\{\frac{d_1}{p},\delta\frac{(d_y-1)}{q} \}), \label{eq:pabrep2}
\ee
\be \|\Delta_{p,q,n,k,\delta}^{xy}\|_1 = 0. \label{eq:pabrep3} \ee
Furthermore, \be J(\ol{f_{\tilde{\bU}_{\bullet}^x,\bU_{\ast_1}^y,
\ldots, \bU_{\ast_{\delta}}^y}})=1 +O(\max
\{\frac{d_x}{p},\frac{d_1}{p},\delta\frac{(d_y-1)}{q} \})
\label{eq:Jbound2} \ee \be
\|\Delta_{p,q,n,k,\delta}^{\tilde{x}y}\|_1 = O\left((d_x/p)\right).
\ee
}

\noindent{\it {Proof:}} We have: \be
\tilde{\mU}^x=(\mU^x(\mU^x)^T)^{-1}\mU^x
\bD_{(\mU^x)^T(\mU^x(\mU^x)^T)^{-2}\mU^x}^{-\frac{1}{2}}. \ee By
block sparsity of $\mathbf \Sigma_{x}, \mU^x$ can be partitioned
as: \be \mU^x = [\underline{\mU}^x, \overline{\mU}^x], \ee where
$\underline{\mU}^x=[\underline{\bU}^x_1,\cdots,\underline{\bU}^x_{d_x}]$
and
$\overline{\mU}^x=[\overline{\bU}^x_1,\cdots,\overline{\bU}^x_{p-d_x}]$
are dependent and independent columns of $\mU^x$, respectively.
Similarly, by block sparsity of $\mathbf \Sigma_{y}$, \be \mU^y =
[\underline{\mU}^y, \overline{\mU}^y], \ee where
$\underline{\mU}^y=[\underline{\bU}^y_1,\cdots,\underline{\bU}^y_{d_y}]$
and
$\overline{\mU}^y=[\overline{\bU}^y_1,\cdots,\overline{\bU}^y_{q-d_y}]$
are dependent and independent columns of $\mU^y$, respectively. By
block sparsity of $\mathbf \Sigma_{xy}$, at
most $d_1$ variables among
$\overline{\bU}^x_1,\cdots,\overline{\bU}^x_{p-d_x}$, are
correlated with columns of $\mU^y$. Assume the correlated
variables are among
$\overline{\bU}^x_1,\cdots,\overline{\bU}^x_{d_2}$. Similarly, at most $d_2$ variables
among $\overline{\bU}^y_1,\cdots,\overline{\bU}^y_{q-d_y}$ are
correlated with columns of $\mU^x$. Without loss of generality,
assume the correlated variables are among
$\overline{\bU}^y_1,\cdots,\overline{\bU}^y_{d_1}$.

The columns of $\overline{\mU}^x$, are i.i.d. and uniform over the
unit sphere $S_{n-2}$. Therefore, as $p \rightarrow \infty$: \be
\frac{1}{p-d_x} \overline{\mU}^x (\overline{\mU}^x)^T \rightarrow
E[\overline{\bU}^x_1 (\overline{\bU}^x_1)^T] = \frac{1}{n-1}
\bI_{n-1}. \ee Also, since the entries of $1/d_x \underline{\mU}^x
(\underline{\mU}^x)^T$ are bounded by one, we have: \be
\frac{1}{p} \underline{\mU}^x (\underline{\mU}^x)^T =
\bO(d_x/p), \ee where $\bO(u)$ is an $(n-1)\times(n-1)$ matrix whose
entries are $O(u)$. Hence: \be (\mU^x (\mU^x)^T)^{-1} \mU^x =
\underline{\mU}^x (\underline{\mU}^x)^T + \overline{\mU}^x
(\overline{\mU}^x)^T \mU^x \nonumber \\ =
\frac{n-1}{p}(\bI_{n-1}+\bO(d_x/p))^{-1} \mU^x \nonumber \\ =
\frac{n-1}{p} \mU^x (1+O(d_x/p)). \label{Mulbounded}\ee Hence, as
$p \rightarrow \infty$: \be (\mU^x)^T (\mU^x (\mU^x)^T)^{-2} \mU^x = \nonumber \\
= (\frac{n-1}{p})^2 (\mU^x)^T \mU^x (1 + O(d_x/p)). \ee Thus: \be
\bD_{(\mU^x)^T (\mU^x (\mU^x)^T)^{-2} \mU^x} = \left(\frac{p}{n-1}
\bI_{n-1}(1+O(d_x/p))\right). \label{Dbounded} \ee Combining
\eqref{Dbounded} and \eqref{Mulbounded} concludes \eqref{tildeZ}.

Now we prove relations \eqref{eq:pabrep2} and \eqref{eq:pabrep3}.
Define the partition ${\mathcal C} = {\mathcal D} \cup {\mathcal
D}^c$ of the index set ${\mathcal C}$ defined in \eqref{indexset},
where ${\mathcal D} = \{\vec{i}=(i_0,i_1,\cdots,i_{\delta}):$
$i_0$ is among $p-d_1$ columns of $\mU^x$ that are uncorrelated of
columns of $\mU^y$ and at most one of $i_1,\cdots,i_{\delta}$ is
less than or equal to $d_y \}$ is the set of $(\delta+1)$-tuples
restricted to columns of $\mU^x$ and $\mU^y$ that are independent.
We have: \be && J(\ol{f_{\bU_{\bullet}^{x}, \bU_{\ast_1}^y, \ldots,
\bU_{\ast_\delta}^y}})= |{\mathcal C}|^{-1}  2^{-\delta} \sum_{s_1, \ldots ,s_{\delta}
\in \{-1,1\}}
\label{eq:Crossavgfubi} \nonumber \\
&&  (\sum_{ \vec{i} \in {\mathcal D}} + \sum_{ \vec{i}
\in {\mathcal D}^c}) J(f_{s_0 \bU_{i_0}^x,s_1 \bU_{i_1}^y,\ldots,
s_{\delta} \bU_{i_\delta}^y}),  \ee and \be
\|\Delta^{xy}_{p,q,n,k,\delta}\|_1 = |{\mathcal C}|^{-1} (\sum_{
\vec{i} \in {\mathcal D}} + \sum_{ \vec{i} \in {\mathcal D}^c})
\Delta_{p,q,n,k,\delta}^{xy}(\vec{i}).\ee But, $ J(f_{s_0
\bU_{i_0}^x,s_1 \bU_{i_1}^y,\ldots, s_{\delta} \bU_{i_\delta}^y})=
1$ for $\vec{i} \in  {\mathcal D}$ and
$\Delta_{p,q,n,k,\delta}^{xy}(\vec{i}) = 0$ for $\vec{i} \in
{\mathcal C}$. Moreover, we have: \be \frac{|{\mathcal
D}|}{|{\mathcal C}|} =
O(\frac{(p-d_1)(q-d_y+1)^\delta}{pq^\delta}). \ee Thus: \be
J(\ol{f_{\bU_{\bullet}^{x}, \bU_{\ast_1}^y, \ldots,
\bU_{\ast_\delta}^y}}) = 1 +O(\max
\{\frac{d_1}{p},\delta\frac{(d_y-1)}{q} \}). \ee  Moreover, since
$\tilde{\mU}^x = \mU^x (1 + O(d_x/p))$, $f_{\tilde{\bU}^x_{i_0},
\bU_{i_1}^y,\ldots, \bU_{i_\delta}^y}=f_{\bU^x_{i_0},
\bU_{i_1}^y,\ldots, \bU_{i_\delta}^y}(1+O(d_x/p))$. This
concludes: \be J(\ol{f_{\tilde{\bU}_{\bullet}^{x}, \bU_{\ast_1}^y,
\ldots, \bU_{\ast_\delta}^y}}) = 1 +O(\max
\{\frac{d_x}{p},\frac{d_1}{p},\delta\frac{(d_y-1)}{q} \}), \ee and
\be \|\Delta^{\tilde{x}y}_{p,q,n,k,\delta}\|_1 = O(d_x/p). \ee \qed

\noindent{\it Proof of Proposition \ref{Prop:UpperBound}:}
First we prove the theorem for $q=1$. Without loss of generality assume \be Y = a_1X_1 + a_2X_2 + \cdots + a_kX_k + \sigma N, \ee where $N$ is follows the standard normal distribution. Note that since $q=1$, $a_1,\cdots,a_k$ are scalars. Defining $\mathbf{b}=\mathbf{\Sigma}^{1/2} \mathbf{a}$, the response $Y$ can be written as:
\be
 Y = a_1Z_1 + a_2Z_2 + \cdots + a_kZ_k + \sigma N, 
\ee
in which $Z_1,\cdots,Z_k$ are i.i.d. standard normal random variables. Assume $\bU_1,\cdots,\bU_p,\bU_N$ represent the U-scores (which are in
$S_{n-2}$) corresponding to $Z_1,\cdots,Z_p,N$, respectively.
It is easy to see:
\begin{equation}
\bU_y = \frac{b_1\bU_1 + b_2\bU_2 + \cdots + b_k\bU_k + \sigma U_N }{\| b_1\bU_1 + b_2\bU_2
+ \cdots + b_k\bU_k + \sigma \bU_N \|}.
\end{equation}
If $\bU$ and $\bV$ are the U-scores corresponding to two random
variables, and $r$ is the correlation coefficient between the two
random variables, we have:
\begin{equation}
|r| = 1 - \frac{(\min \{ \|\bU-\bV\|, \|\bU+\bV\|\})^2}{2}.
\end{equation}
Let $r_{y,i}$ represent the sample correlation between $Y$ and $X_i$. Here, we want to upper bound $\text{prob}\{ |r_{y,1}| < |r_{y,k+1}|\}$. We
have:
\be
\text{prob}\{ |r_{y,1}| < |r_{y,k+1}|\} \nonumber &=& \\ \nonumber
\text{prob}\{1 -\frac{(\min \{ \|\bU_1-\bU_y\|, \|\bU_1+\bU_y\|\})^2}{2} &<& \nonumber \\
 1 - \frac{(\min \{ \|\bU_{k+1}-\bU_y\|, \|\bU_{k+1}+\bU_y\|\})^2}{2}\} &=& \\
\nonumber  \text{prob}\{\min \{ \|\bU_1-\bU_y\|, \|\bU_1+\bU_y\|\} &>&
\nonumber \\  \min \{ \|\bU_{k+1}-\bU_y\|, \|\bU_{k+1}+\bU_y\|\}\} &\leq& \\ \nonumber  \text{prob}\{\|\bU_1-\bU_y\| &>& \nonumber \\
 \min \{ \|\bU_{k+1}-\bU_y\|, \|\bU_{k+1}+\bU_y\|\}\} &=& \nonumber \\   \text{prob}\{ \{\|\bU_1-\bU_y\|
> \|\bU_{k+1}-\bU_y\|\} &\cup& \nonumber \\    \{\|\bU_1-\bU_y\| > \|\bU_{k+1}+\bU_y\|\} \} &\leq& \nonumber \\   \text{prob}\{\|\bU_1-\bU_y\|
> \|\bU_{k+1}-\bU_y\|\} &+& \nonumber  \\   \text{prob} \{\|\bU_1-\bU_y\| > \|\bU_{k+1}+\bU_y\|\} &=& \\
 2 ~\text{prob}\{\|\bU_1-\bU_y\|> \|\bU_{k+1}-\bU_y\|\},
\label{p1def}
\ee
in which, the last inequality holds since $\bU_{k+1}$ is uniform over
$S_{n-2}$ and is independent of $\bU_1$ and $\bU_y$. Therefore, it
suffices to upper bound $p_1:=\text{prob}\{\|\bU_1-\bU_y\|>
\|\bU_{k+1}-\bU_y\|\}$. Define:
\begin{equation}
\bV = b_2 \bU_2 + \cdots + b_k \bU_k,
\end{equation}
and
\begin{equation}
\bU_* = \bV/\|\bV\|.
\end{equation}
By symmetry, $\bU_*$ is uniform over $S_{n-2}$. Hence:
\begin{equation}
\bU_y=\frac{b_1 \bU_1+\|\bV\|\bU_*}{\|b_1 \bU_1+\|\bV\|\bU_*\|}.
\end{equation}
Since $\|\bV\| \leq |b_2|+\cdots+|b_k|$, we have:
\begin{equation}
\frac{|b_1|}{\|\bV\|} \geq
\frac{|b_1|}{|b_2|+\cdots+|b_k|}=\frac{|b_1|}{c_1},
\end{equation}
where $c_1:=|b_2|+\cdots+|b_k|$. Define:
\begin{equation}
\theta_1 = \cos^{-1}(\bU_y^T \bU_1),
\end{equation}
and
\begin{equation}
\theta_1 = \cos^{-1}(\bU_y^T \bU_*).
\end{equation}
It is easy to see that:
\begin{equation}
\frac{\sin \theta_1}{\sin \theta_2} \leq \frac{c_1}{|b_1|}.
\end{equation}
For each $0 \leq \theta \leq \pi$, define:
\begin{equation}
\beta_1(\theta) = \max_{0 \leq \theta' \leq \pi}
\frac{\theta}{\theta+\theta'} ~~\text{s.t.}~~ \frac{\sin
\theta}{\sin \theta'} \leq \frac{c_1}{|b_1|}.
\end{equation}
Now fix the point $\bU_1$ on $S_{n-2}$. Define $f(\theta)$ as the
probability distribution of $\theta_2$. Also, define $p(\theta)$
as the probability that the angle between the uniformly
distributed (over $S_{n-2}$) point $\bU_{k+1}$ and $\bU_y$ is less
than $\theta$. Since $\bU_1$ is independent of $\bU_*$ and $\bU_{k+1}$
is independent of $\bU_y$, clearly:
\begin{equation}
p(\theta)=\int_{0}^{\theta} f(\theta') d\theta'.
\label{eq:cap_prob}
\end{equation}
We have:
\begin{eqnarray}
p_1 \leq \int_{0}^{\pi} p(\beta_1(\theta)\theta) f(\theta) d\theta
\nonumber \\ = \int_{0}^{\pi/2} \left( p(\beta_1(\theta)\theta)+p(\beta_1(\pi
- \theta)(\pi - \theta)) \right) f(\theta)d\theta,
\end{eqnarray}
where the last equality holds because $f(\theta)=f(\pi - \theta)$.
Noting the fact that:
\begin{eqnarray}
\int_{0}^{\pi} p(\theta) f(\theta) d\theta
= \int_{0}^{\pi/2}(p(\theta)+ \nonumber \\ p(\pi - \theta))f(\theta)d\theta
 = \int_{0}^{\pi/2}f(\theta) d\theta = \frac{1}{2}.
\end{eqnarray}
we conclude:
\begin{eqnarray}
p_1 \leq \frac{1}{2} -
\int_{0}^{\pi/2}\{(p(\theta)-p(\beta_1(\theta)\theta))+ \nonumber \\ (p(\pi-\theta)-p(\beta_1(\pi
- \theta)(\pi - \theta)))\}f(\theta)d\theta.
\end{eqnarray}
Hence by \eqref{eq:cap_prob}, for any $0 < \theta_0 < \pi/2$:
\begin{equation}
p_1 \leq \frac{1}{2} - \int_{\theta_0}^{\pi/2} p_{\gamma_1}(\theta)
f(\theta)d\theta,
\end{equation}
in which
\be
p_{\gamma_1}(\theta) &=& p(\theta + \gamma_1\theta)-p(\theta - \gamma_1 \theta)
\nonumber \\ &=& \text{prob}\{ \theta - \gamma_1 \theta \leq
\theta_2 \leq \theta + \gamma_1\theta\},
\ee
with
\begin{equation}
\gamma_1 = \min_{\theta_0 \leq \theta \leq \pi-\theta_0} 1-\beta_1(\theta) =
1 - \max_{\theta_0 \leq \theta \leq \pi - \theta_0} \beta_1(\theta).
\end{equation}
It is easy to check that $\gamma_1 > 0$. Therefore, since
$p_{\gamma_1}(\theta)$ is an increasing functions of $\theta$ for $0
\leq \theta \leq \pi/2$, we conclude:
\begin{equation}
p_1 \leq \frac{1}{2} - \int_{\theta_0}^{\pi/2}
p_{\gamma_1}(\theta_0) f(\theta)d\theta.
\end{equation}
Choose $\theta_0$ so that $\theta_0=\frac{\pi}{2+\gamma_1}$. We have:
\begin{eqnarray}
p_1 \leq \frac{1}{2} - p_{\gamma_1}(\pi/(2+\gamma_1))
\int_{\pi/(2+\gamma_1)}^{\pi/2} f(\theta)d\theta \nonumber \\
= \frac{1}{2} -
\int_{\pi(1-\gamma_1)/(2+\gamma_1)}^{\pi(1+\gamma_1)/(2+\gamma_1)}
f(\theta)d\theta \int_{\pi/(2+\gamma_1)}^{\pi/2} f(\theta)d\theta
\nonumber \\ \leq \frac{1}{2} - \int_{\pi/2 -\gamma_1 \pi/6}^{\pi/2
+\gamma_1 \pi/6} f(\theta)d\theta \int_{\pi/2 -\gamma_1 \pi/6}^{\pi/2} f(\theta)d\theta
\nonumber \\ \leq \frac{1}{2} - 2 \left(\int_{\pi/2 -\gamma_1
\pi/6}^{\pi/2} f(\theta)d\theta \right)^2,
\label{eq:p1bound}
\end{eqnarray}
in which, the last inequality holds, since $0 < \gamma_1 < 1$.
Defining $\lambda_1 = \sin(\pi/2 -\gamma_1 \pi/6)$ and using the
formula for the area of the spherical cap, we will have:
\be
&&\int_{\pi/2 -\gamma_1 \pi/6}^{\pi/2} f(\theta)d\theta = \nonumber \\
&&\frac{I_1((n-2)/2,1/2)-I_{\lambda_1}((n-2)/2,1/2)}{2I_1((n-2)/2,1/2)},
\ee
in which
\begin{equation}
I_x(a,b)=\frac{\int_{0}^{x}t^{a-1}(1-t)^{b-1}dt}{\int_{0}^{1}t^{a-1}(1-t)^{b-1}dt},
\end{equation}
is the regularized incomplete beta function. Hence:
\begin{equation}
\int_{\pi/2 -\gamma_1 \pi/6}^{\pi/2} f(\theta)d\theta =
\frac{\int_{\lambda_1}^{1}t^{(n-4)/2}/\sqrt{1-t}dt}{2\int_{0}^{1}t^{(n-4)/2}/\sqrt{1-t}dt}.
\end{equation}
Note that we have:
\begin{eqnarray}
\frac{\int_{\lambda_1}^{1}t^{(n-4)/2}/\sqrt{1-t}dt}{2\int_{0}^{\lambda_1}t^{(n-4)/2}/\sqrt{1-t}dt}
\nonumber \\ \geq
\frac{\int_{\lambda_1}^{1}t^{(n-4)/2}/\sqrt{1-\lambda_1}dt}{2\int_{0}^{1}t^{(n-4)/2}/\sqrt{1-\lambda_1}dt}
\nonumber \\ = \frac{1 - \lambda_1^{(n-2)/2}}{\lambda_1^{(n-2)/2}}
:=\kappa_1.
\end{eqnarray}
Hence:
\be
\int_{\pi/2 -\gamma_1 \pi/6}^{\pi/2} f(\theta)d\theta  &\geq&
\frac{\int_{\lambda_1}^{1}t^{(n-4)/2}/\sqrt{1-t}dt}{2(1+1/\kappa_1)\int_{\lambda_1}^{1}t^{(n-4)/2}/\sqrt{1-t}dt}\nonumber
\\ &=& \frac{\kappa_1}{2(\kappa_1+1)} = \frac{1-\lambda_1^{(n-2)/2}}{2}.
\ee
Hence by \eqref{eq:p1bound}:
\begin{equation}
p_1 \leq \lambda_1^{(n-2)/2} - \lambda_1^{n-2} \leq \lambda_1^{(n-2)/2}.
\end{equation}
Therefore, $p_1$ decreases at least exponentially by $n$.

Assume $P(i)$ for $1 \leq i \leq k$, represents the probability
that the active variable $X_i$ is not among the selected $k$
variables. By \eqref{p1def} and using the union bound we have:
\begin{equation}
P(1) \leq 2(p-k)\lambda_1^{(n-2)/2}.   \label{PermUP}
\end{equation}
Similar inequalities can be obtained for $P(2),\cdots,P(k)$ which
depend on $\lambda_2,\cdots,\lambda_k$, respectively. Finally,
using the union bound, the probability $P$ that all the active
variables are correctly selected satisfies:
\begin{equation}
P \geq 1 - 2(p-k)\sum_{i=1}^{k}\lambda_i^{(n-2)/2} \geq 1 - 2k(p-k)\lambda^{(n-2)/2},
\label{ExactUP}
\end{equation}
where $\lambda:= \max_{1 \leq i \leq k}\lambda_i$. This concludes that if $n = \Theta(\log p)$, with probability at least $1-1/p$ the exact support can be recovered using PCS.

For $q>1$, by union bound, the probability of error becomes at most $q$ times larger and this concludes the statement of proposition~\ref{Prop:UpperBound}. \qed

\noindent{\it Proof of Proposition \ref{Prop:UpperBound2}:} First we consider a two-stage predictor similar to the one introduced in previous section with the difference that the $n$ samples which are used in stage $1$ are not used in stage $2$. Therefore, there are $n$ and $t-n$ samples used in the first and the second stages, respectively. Following the notation introduced in previous section, we represent this two-stage predictor by $n|(t-n)$.  The asymptotic results for the $n|(t-n)$ two-stage predictor will be shown to hold as well for the $n|t$ two-stage predictor.

Using inequalities of the form \eqref{PermUP} and the union bound, it is straightforward to see that for any subset $\pi \neq \pi_0$ of $k$ elements of $\{1,\cdots,p \}$, the probability that $\pi$ is the outcome of variable selection via PCS, is
bounded above by $2k(p-k)c_{\pi}^n$, in which $0 < c_{\pi}<1$ is a constant that depends on the quantity
\begin{equation}
\min_{j \in \pi_0 \cap \pi^c} \frac{|a_j|}{\sum_{l \in \pi_0} |a_{l}|}.
\end{equation}
The expected MSE of the $n|(t-n)$ algorithm can be written as:
\begin{equation}
\text{E}[\text{MSE}] = \sum_{\pi \in S_k^p, \pi \neq \pi_0} p(\pi)\text{E}[\text{MSE}_{\pi}] +
p(\pi_0)\text{E}[\text{MSE}_{\pi_0}],
\end{equation}
where $S_k^p$ is the set of all $k$-subsets of $\{1,\cdots,p\}$, $p(\pi)$ is the probability that the outcome of variable
selection via PCS is the subset $\pi$, and $\text{MSE}_{\pi}$ is the MSE of OLS  stage when the indices of the selected variables are the elements of $\pi$.
Therefore using the bound \eqref{ExactUP}, the expected MSE is upper bounded as below:
\begin{eqnarray}
\text{E}[\text{MSE}] \leq 2k(p-k)\sum_{\pi \in S_k^p, \pi \neq \pi_0}
 c_{\pi}^n \text{E}[\text{MSE}_{\pi}] + \nonumber \\ (1-2k(p-k)c_0^n)\text{E}[\text{MSE}_{\pi}],
\end{eqnarray}
$c_0$ is a constant that depends on the quantity \eqref{quantity1}. It can be shown that if there is at least one wrong variable
selected ($\pi \neq \pi_0$), the OLS estimator is biased and the expected MSE
converges to a positive constant $M_{\pi}$ as $(t-n) \rightarrow
\infty$. When all the variables are selected correctly (subset
$\pi_0$), MSE goes to zero with rate $O(1/(t-n))$. Hence:
\be
\text{E}[\text{MSE}] \leq \nonumber 2k(p-k)\sum_{\pi \in S_k^p, \pi \neq \pi_0} c_{\pi}^n
M_{\pi} + \nonumber \\  (1-2k(p-k)c_0^n) O(1/(t-n)) \leq \nonumber \\ 2k(p-k)C_1 C^n + (1-2k(p-k) C^n) C_2/(t-n) \label{upper_objective},
\ee
where $C,C_1$ and $C_2$ are constants that do not depend on $n$ or $p$
but depend on the quantities $\sum_{j \in \pi_0} a_{j}^2$ and $\min_{j \in \pi_0} |a_{j}| /\sum_{l \in \pi_0} |a_{l}|$.

On the other hand since at most $t$ variables could be used in OLS
stage, the expected MSE is lower bounded:
\begin{equation}
\text{E}[\text{MSE}] \geq \Theta (1/t).
\label{eq:lowerb}
\end{equation}



It can be seen that the minimum of \eqref{upper_objective} as a
function of $n$, subject to the constraint \eqref{eq:assaycostcond}, happens for
$n=O(\log t)$ if $\Theta(\log t) \leq \frac{\mu - tk}{p-k}$; otherwise
it happens for 0. 
If $ \Theta(\log t) \leq \frac{\mu - tk}{p-k}$, the minimum value
attained by the upper bound \eqref{upper_objective} is $\Theta(1/t)$ which is
as low as the lower bound \eqref{eq:lowerb}. This shows that for large $t$, the optimal number of samples that should be assigned to the PCS stage of the $n|(t-n)$ predictor is $n=O(\log t)$. As $t
\rightarrow \infty$, since $n=O(\log t)$, the MSE of the $n|t$ predictor proposed in
Sec. \ref{sec:two-stage} converges to the MSE of the $n|(t-n)$ predictor.
Therefore, as $t \rightarrow \infty$, $n=O(\log t)$ becomes optimal for the $n|t$ predictor as well. \qed

\end{document}